\documentclass[letterpaper, 10 pt, conference]{ieeeconf} 

\IEEEoverridecommandlockouts                              

\usepackage{graphics} 
\usepackage{epsfig} 
\usepackage{mathptmx} 
\usepackage{times} 
\usepackage{amsmath} 
\usepackage{amssymb}  
\usepackage{xcolor}
\usepackage{algorithm2e}
\usepackage{eso-pic}
\makeatletter
\renewcommand{\@algocf@capt@plain}{above}
\renewcommand{\algocf@caption@plain}{\box\algocf@capbox\vskip\AlCapSkip}%
\makeatother

\usepackage{balance}
\usepackage[utf8]{inputenc}
\usepackage[pdfauthor={Bonetto et al.},
            pdftitle={},
            pdfsubject={Synthetic based training},
            pdfkeywords={dataset, synthetic-data, zebra detection, drone imagery, synthetic training, yolo, GRADE, sim2real}]{hyperref}
\let\labelindent\relax
\usepackage{enumitem}
\usepackage{subcaption}
\usepackage{cuted}
\usepackage{multirow}
\hypersetup{
    colorlinks=true,
    linkcolor=blue,
    filecolor=magenta,      
    urlcolor=blue,
}

\usepackage[font=footnotesize,labelfont=bf]{caption}
\usepackage{tikz}

\usepackage[textsize=tiny]{todonotes}

\newcommand{\uproman}[1]{\uppercase\expandafter{\romannumeral#1}}

\usepackage{soul}

\title{\LARGE\bf{Synthetic Data-based Detection of Zebras in Drone Imagery}}

\author{Elia Bonetto$^{*,\dagger}$ ~\IEEEmembership{Student Member,~IEEE,} and Aamir Ahmad$^{\dagger,*}$~\IEEEmembership{Senior Member,~IEEE}
    \thanks{$^*$Max Planck Institute for Intelligent Systems, Tübingen, Germany. {\tt\footnotesize {firstname.lastname}@tuebingen.mpg.de}}
    \thanks{$^\dagger$Institute of Flight Mechanics and \mbox{Controls}, University of Stuttgart,  Germany. {\tt\footnotesize {firstname.lastname}@ifr.uni-stuttgart.de}}
    \thanks{The authors thank Eric Price, Nitin Saini, Egor Iuganov, Chenghao Xu, and Benedikt Schwämmle for their help in collecting and processing the data.}
    \thanks{We would like to extend our deepest gratitude to the Wilhelma Zoo in Stuttgart, and in particular to Ms. Ulrike Rademacher, for permitting us to record videos of Grévy's zebras on the premises of the Wilhelma Zoo.}
    \thanks{The authors thank the International Max Planck Research School for Intelligent Systems (IMPRS-IS) for supporting Elia Bonetto.}
    \thanks{979-8-3503-0704-7/23/\$31.00 ©2023 IEEE} 
}

\begin{document}

\AddToShipoutPictureBG*{%
  \AtPageUpperLeft{%
    \setlength\unitlength{1in}%
    \hspace*{\dimexpr0.5\paperwidth\relax}
    \makebox(0,-0.75)[c]{\parbox{0.8\textwidth}{\centering This paper has been accepted for publication in the \\ 2023 IEEE European Conference on Mobile Robots.\\
    Please cite as: Bonetto, E. and Ahmad, A. (2023). Synthetic Data-based Detection of Zebras in Drone Imagery. \textit{IEEE European Conference on Mobile Robots (ECMR)}.}}%
}}

	\maketitle
\begin{abstract}
Nowadays, there is a wide availability of datasets that enable the training of common object detectors or human detectors. These come in the form of labelled real-world images and require either a significant amount of human effort, with a high probability of errors such as missing labels, or very constrained scenarios, e.g. VICON systems. On the other hand, uncommon scenarios, like aerial views, animals, like wild zebras, or difficult-to-obtain information, such as human shapes, are hardly available. To overcome this, synthetic data generation with realistic rendering technologies has recently gained traction and advanced research areas such as target tracking and human pose estimation. However, subjects such as wild animals are still usually not well represented in such datasets. In this work, we first show that a pre-trained YOLO detector can not identify zebras in real images recorded from aerial viewpoints. To solve this, we present an approach for training an animal detector using only synthetic data. We start by generating a novel synthetic zebra dataset using GRADE, a state-of-the-art framework for data generation. The dataset includes RGB, depth, skeletal joint locations, pose, shape and instance segmentations for each subject. We use this to train a YOLO detector from scratch. Through extensive evaluations of our model with real-world data from i) limited datasets available on the internet and ii) a new one collected and manually labelled by us, we show that we can detect zebras by using only synthetic data during training. The code, results, trained models, and both the generated and training data are provided as open-source at~\url{https://eliabntt.github.io/grade-rr}.
\end{abstract}


	\section{INTRODUCTION}
\label{sec:intro}

\begin{figure*}[!ht]
  \includegraphics[width=0.245\textwidth, height=2.5cm]{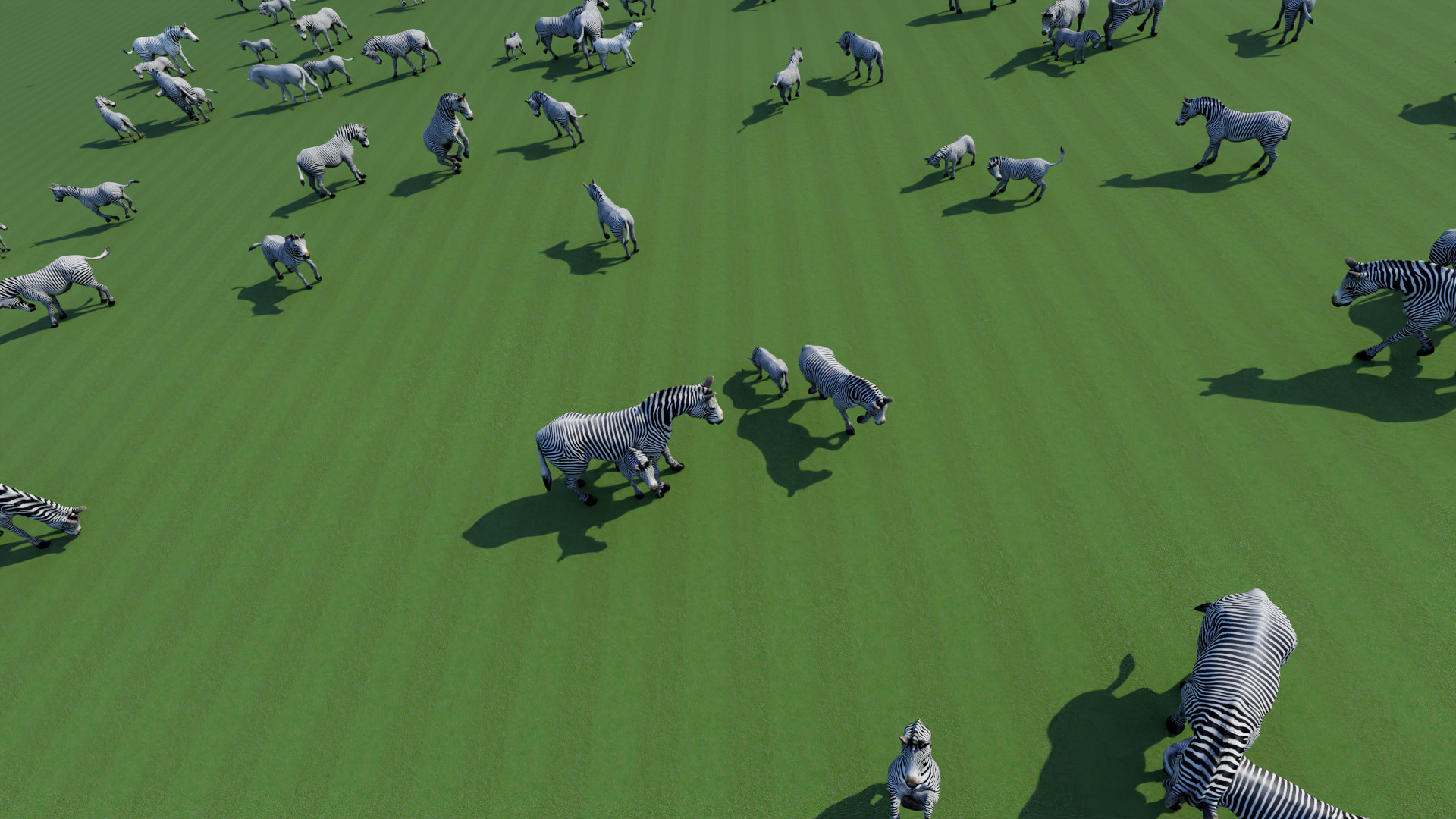}   \hfill
  \includegraphics[width=0.245\textwidth, height=2.5cm]{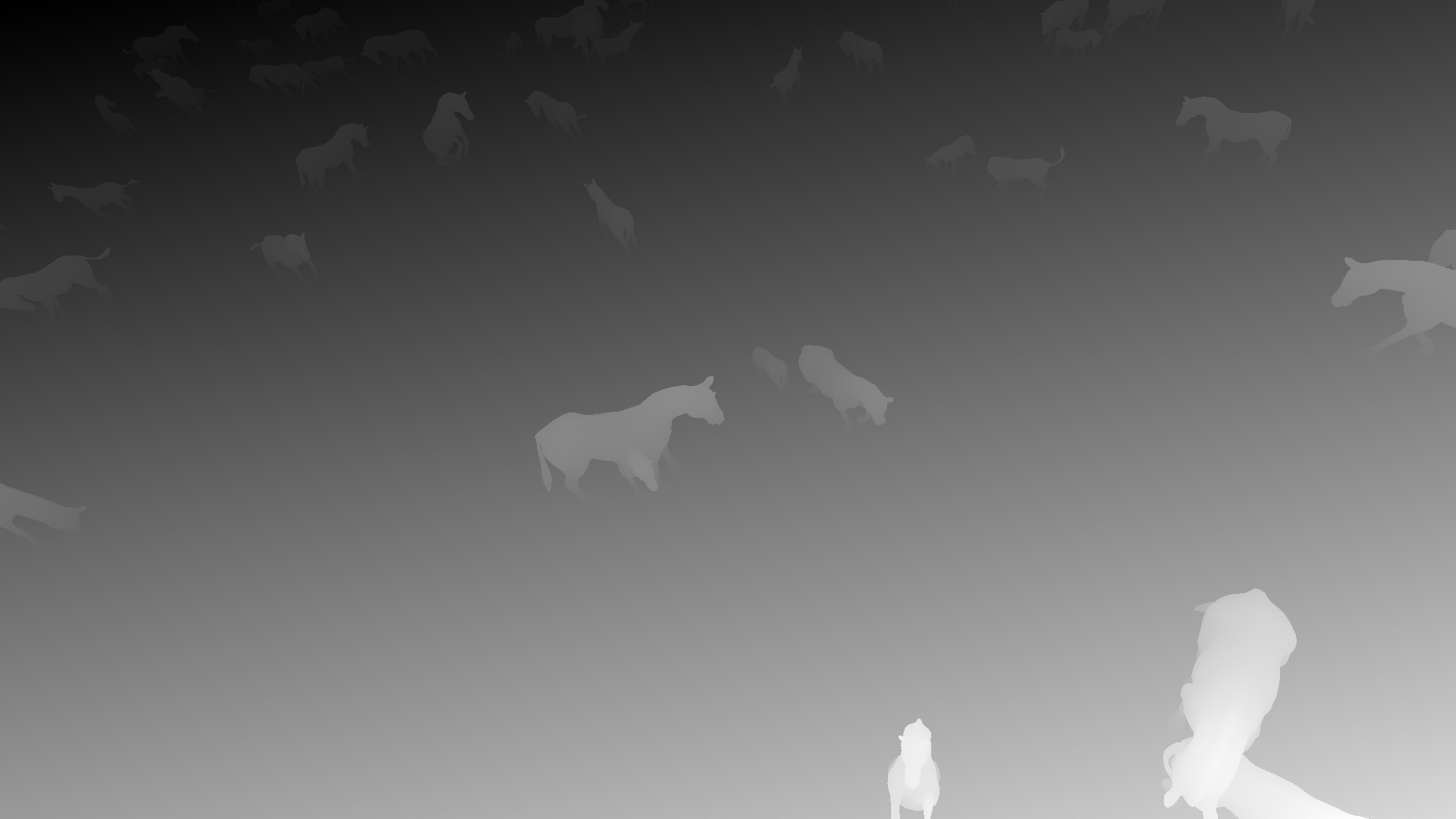}  \hfill
  \includegraphics[width=0.245\textwidth, height=2.5cm]{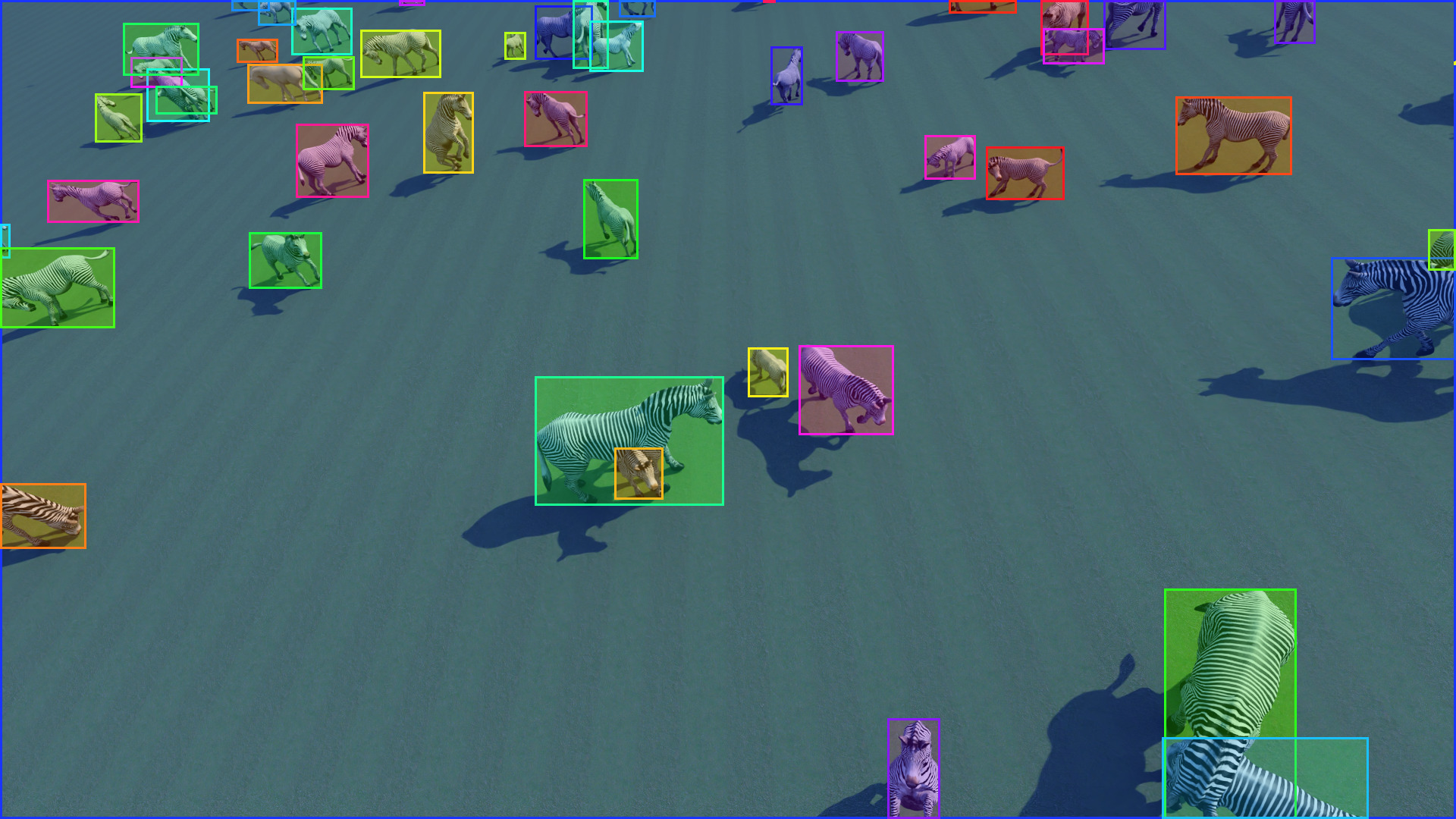} \hfill
  \includegraphics[width=0.245\textwidth, height=2.5cm]{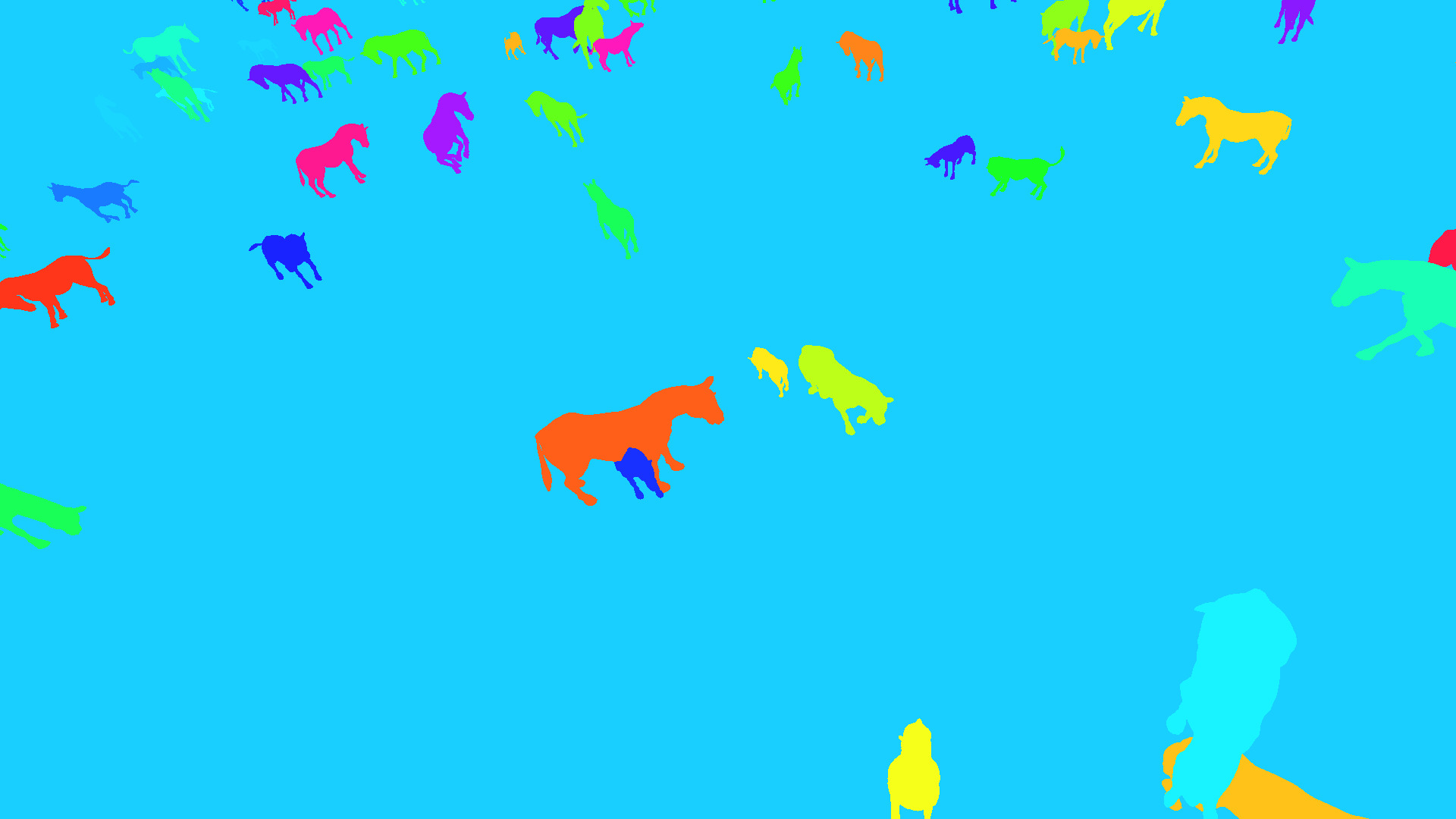} \hfill\\

  \includegraphics[width=0.245\textwidth, height=2.5cm]{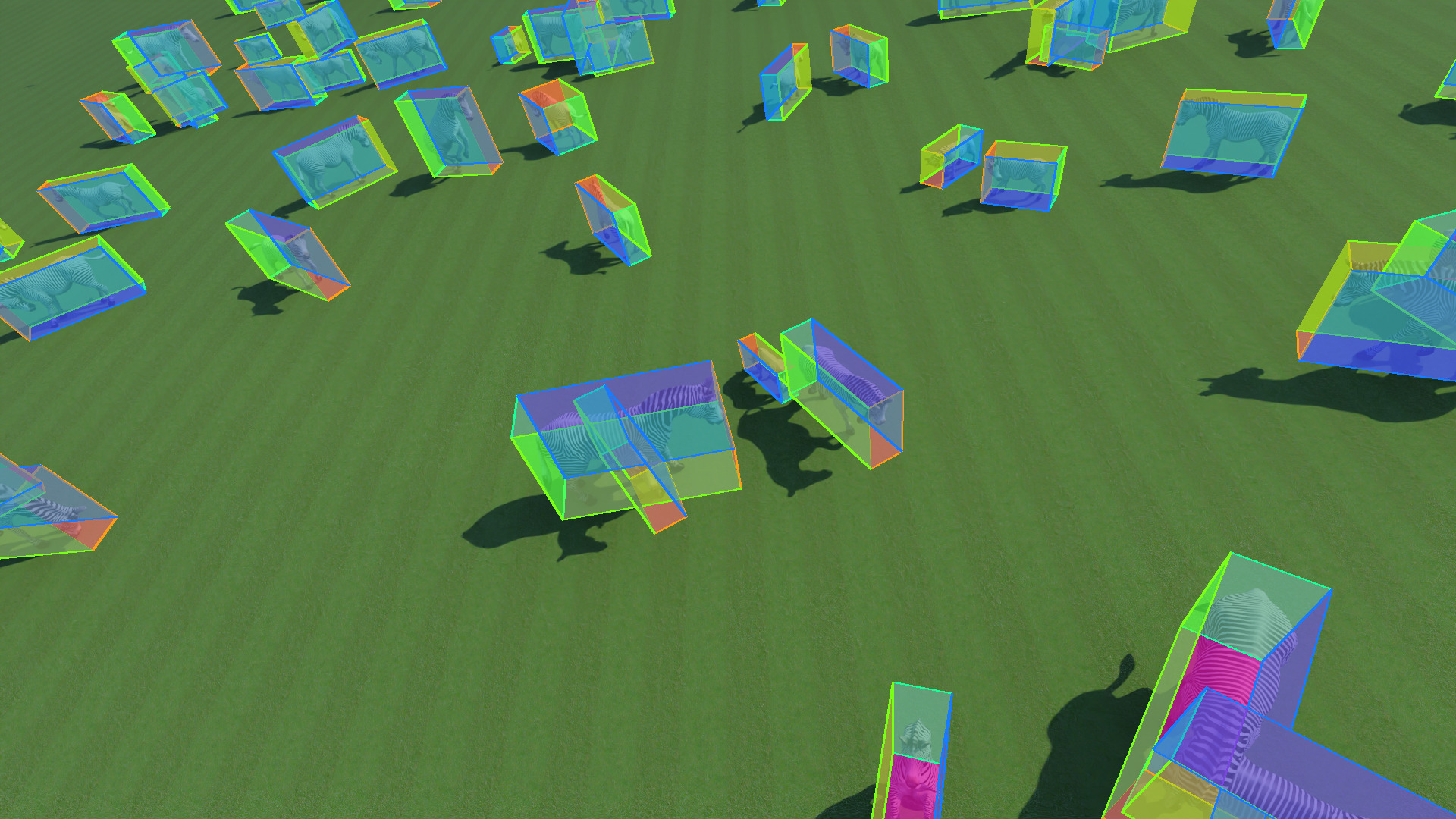}   \hfill
  \includegraphics[width=0.245\textwidth, height=2.5cm]{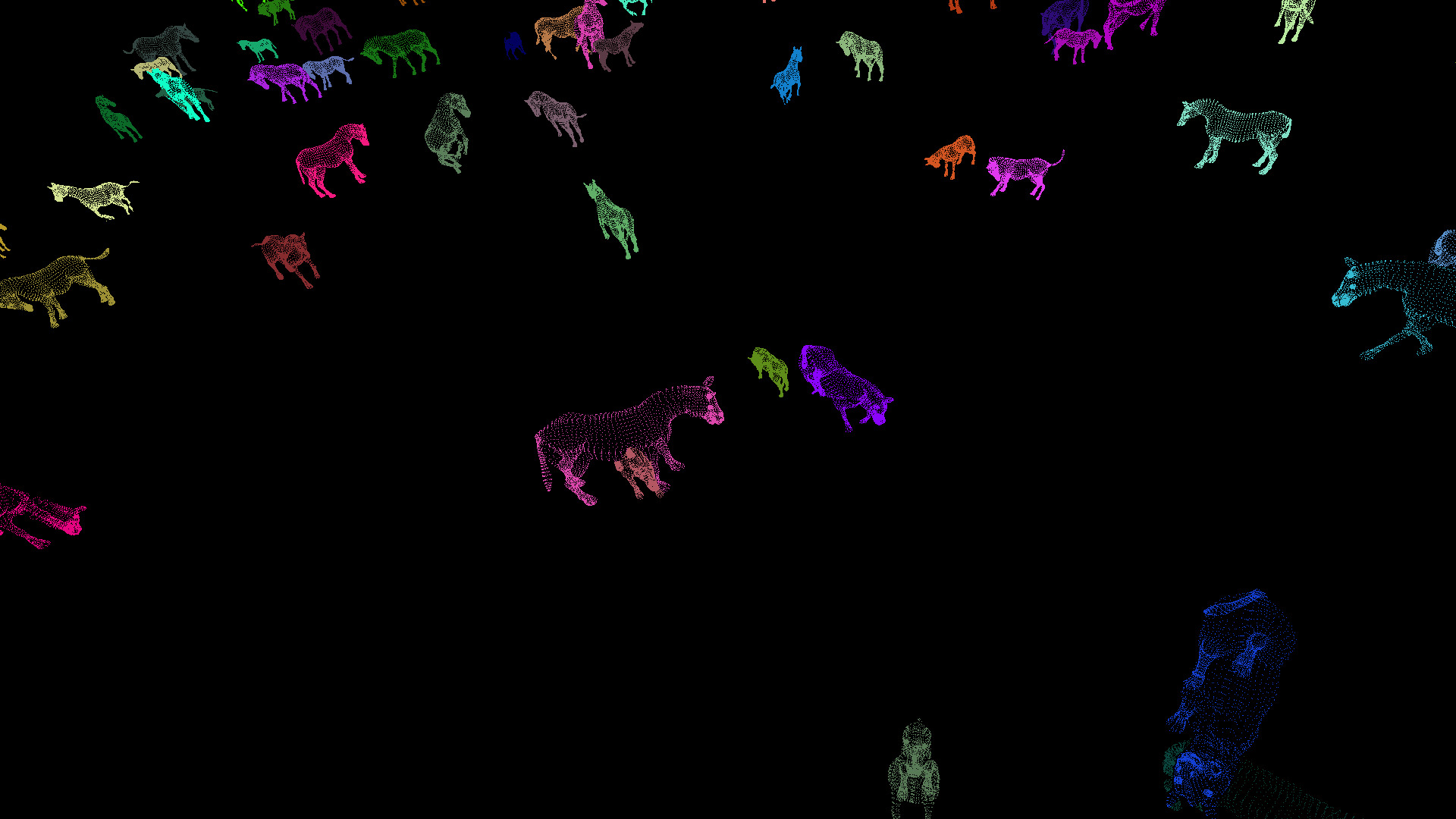} \hfill 
  \includegraphics[width=0.245\textwidth, height=2.5cm]{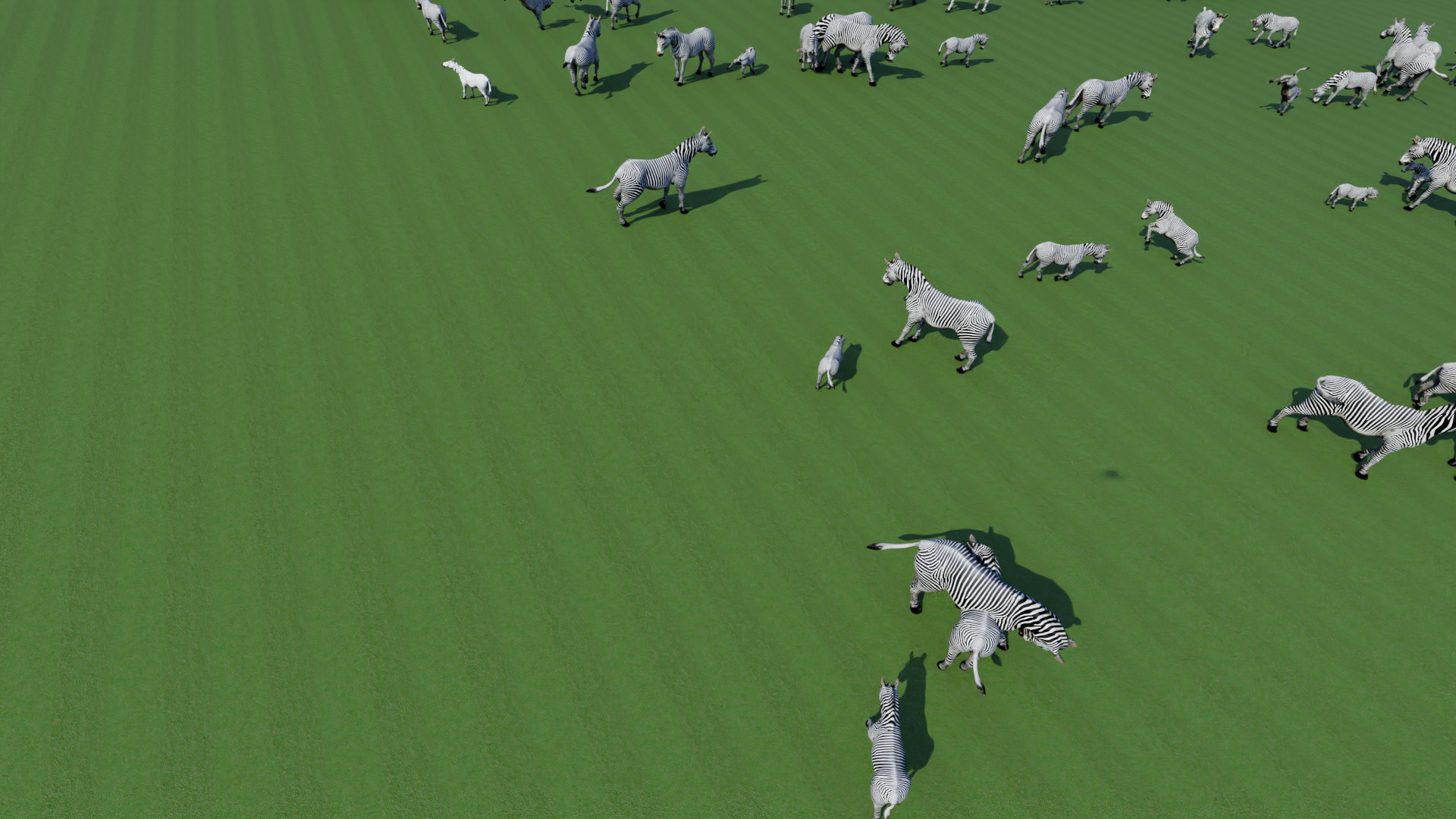}  \hfill
  \includegraphics[width=0.245\textwidth, height=2.5cm]{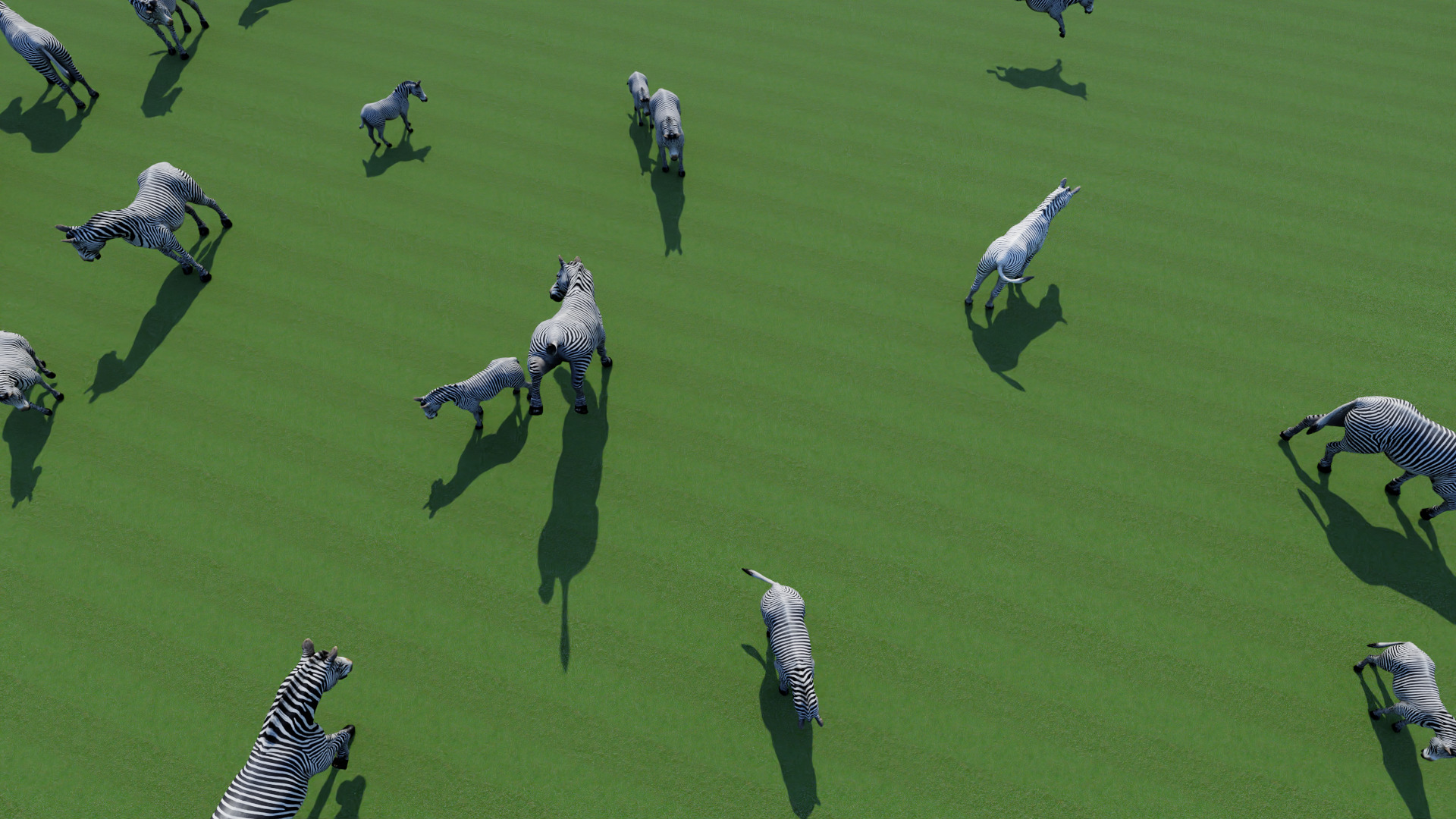} \hfill\\
  
  \caption{An example of the generated data following the pipeline described in Sec.~\ref{sec:gen}. On the first row, from the left, we can see the rendered RGB image, depth data, 2D bounding boxes, and semantic instances. On the second row, from the left, we can see 3D-oriented bounding boxes, the vertices of each mesh drawed over a black background, and the second and third views of the same scene as taken from the other drones. Best viewed in color.}
  \label{fig:gendata}
\end{figure*}

A large dataset that includes realism and diversity in features is a fundamental building block for obtaining any working and reliable deep-learning model. This is especially true when dealing with visual tasks such as detection, semantic segmentation and shape estimation. For these, variability in both visual appearances and environmental conditions as well as a high number of instances are required. A variety of datasets have been introduced during the last decades to address various image-based tasks like MNIST~\cite{lecun-mnisthandwrittendigit-2010}, COCO~\cite{cocodataset}, and PASCAL-VOC~\cite{Everingham10}. These have historically been based on real-world data, be this either images or videos, manually labelled by humans. Apart from being time-consuming and costly, this introduces errors such as missing and wrong labels~\cite{rottmann2023automated}. Examples of these can be visualized in Fig.~\ref{fig:bad-dpk} and Fig.~\ref{fig:bad-coco}. Furthermore, ground truth for specific tasks may not be available either because it is hard to obtain, e.g., shape or skeletal information, or because it requires costly manual labelling procedures. These limitations impede the usage of these datasets in problems such as aerial human pose estimation~\cite{airpose}, animal pose estimation~\cite{SMAL,SMALR}, or aerial wild-animal detection. For these reasons, methodologies to generate synthetic data became more ubiquitous since the advent of rendering engines such as Unity, Blender, Unreal Engine and IsaacSim. These are advantageous in multiple aspects since they allow generation and automatic labelling of ground truth data through full customization possibilities and with minimum human effort~\cite{GRADE}. Indeed, synthetic data has been used in a variety of tasks such as human detection~\cite{GRADE,peoplesanspeople}, pose and shape estimation of humans~\cite{airpose}, and semantic segmentation~\cite{sankaranarayanan2018learning}. However, they usually lack the visual realism necessary to generalize well to real-world data if used alone. Thus, a combination of real and synthetic data is often utilized~\cite{GRADE,peoplesanspeople}. 
\begin{figure}[!t]
    \centering
    \includegraphics[width=\columnwidth]{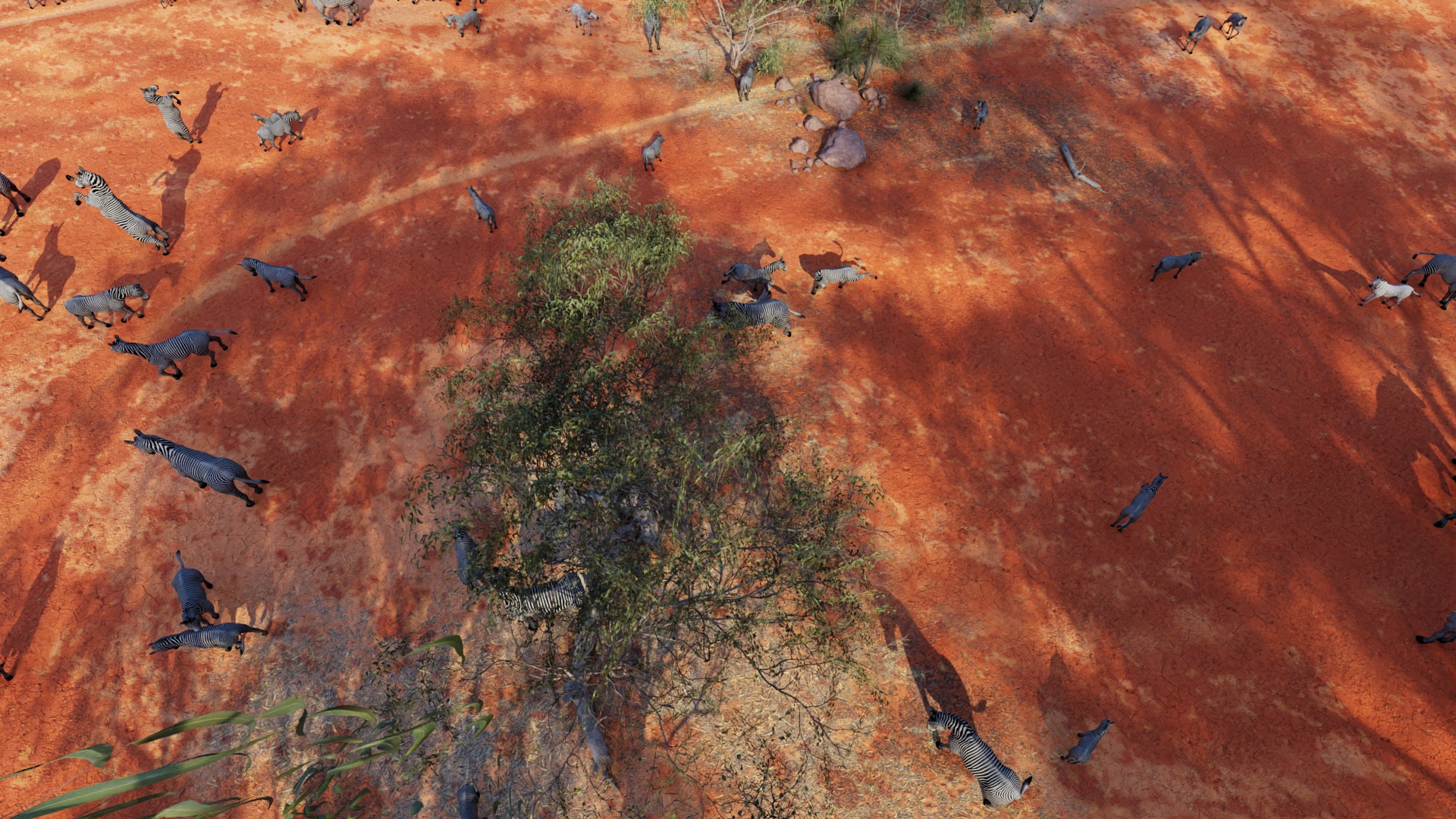}
    \caption{An example image of our synthetically generated zebras in a Savanna environment.}
    \label{fig:cover}
\end{figure}
Moreover, these datasets are usually application-specific and hardly generalize to different scenarios, tasks, or data. For example, wild animals are widely under-represented in datasets such as COCO or PASCAL-VOC~\cite{animalpose,deeplabcut,ap-10k}. Indeed, apart from a limited number of labelled images and videos of uncommon animal species such as zebras, hippopotami, and giraffes, there is also a general lack of variety of scenarios in which those are recorded. Taking zebras as an example, there are only 1916 training and 85 validation images containing at least one instance of them in the COCO dataset. To solve this problem, we generate a new synthetic dataset using our GRADE~\cite{GRADE} framework, a publicly available animated zebra model and environments from the Unreal Engine marketplace. An example of the generated data can be seen in Fig.~\ref{fig:cover} and Fig.~\ref{fig:gendata}. We use this data to train a YOLO-based detector and perform evaluations with an extensive dataset of zebras captured by drones consisting of 104K images and the APT-36K~\cite{apt-36k} dataset. With this, we show that training with our synthetic data outperforms the baseline models trained on real-world datasets. In this paper, we take zebras as an example as it is an endangered species, which is greatly under-represented in currently available datasets. However, the proposed method can be generalized to other animals as well.

The rest of the paper is structured as follows. In Sec.~\ref{sec:soa} we review the current state-of-the-art in simulated worlds and animal datasets. Our method is described thoroughly in Sec.~\ref{sec:approach}. In there, we give a general overview of the system and explain how we generated our data. We then present the results of our experiments in Sec.~\ref{sec:exp} and conclude the work with Sec.~\ref{sec:conc} with comments on known limitations and possible future work.
	\section{Related Work}
\label{sec:soa}
In this section, we focus on two areas: animal-based datasets, and simulation engines. 

\begin{figure}
    \centering
    \includegraphics[width=0.48\columnwidth]{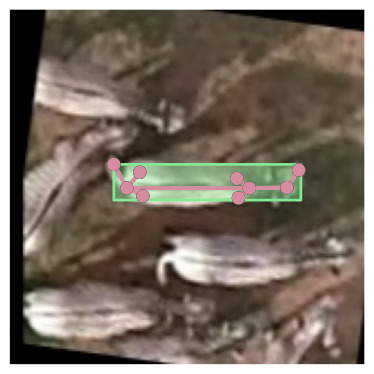}
    \includegraphics[width=0.48\columnwidth]{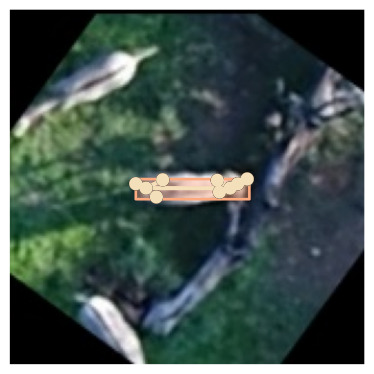}
    \caption{Examples of missing bounding boxes and keypoints from the Grévy's zebra~\cite{dpk-zebrads} dataset. Image ids 869 (left) and 882 (right). }
    \label{fig:bad-dpk}
\end{figure}

\begin{figure}
    \centering
      \subcaptionbox{ID: 20164, missing bounding boxes}{\includegraphics[width=0.495\columnwidth, height=3.5cm]{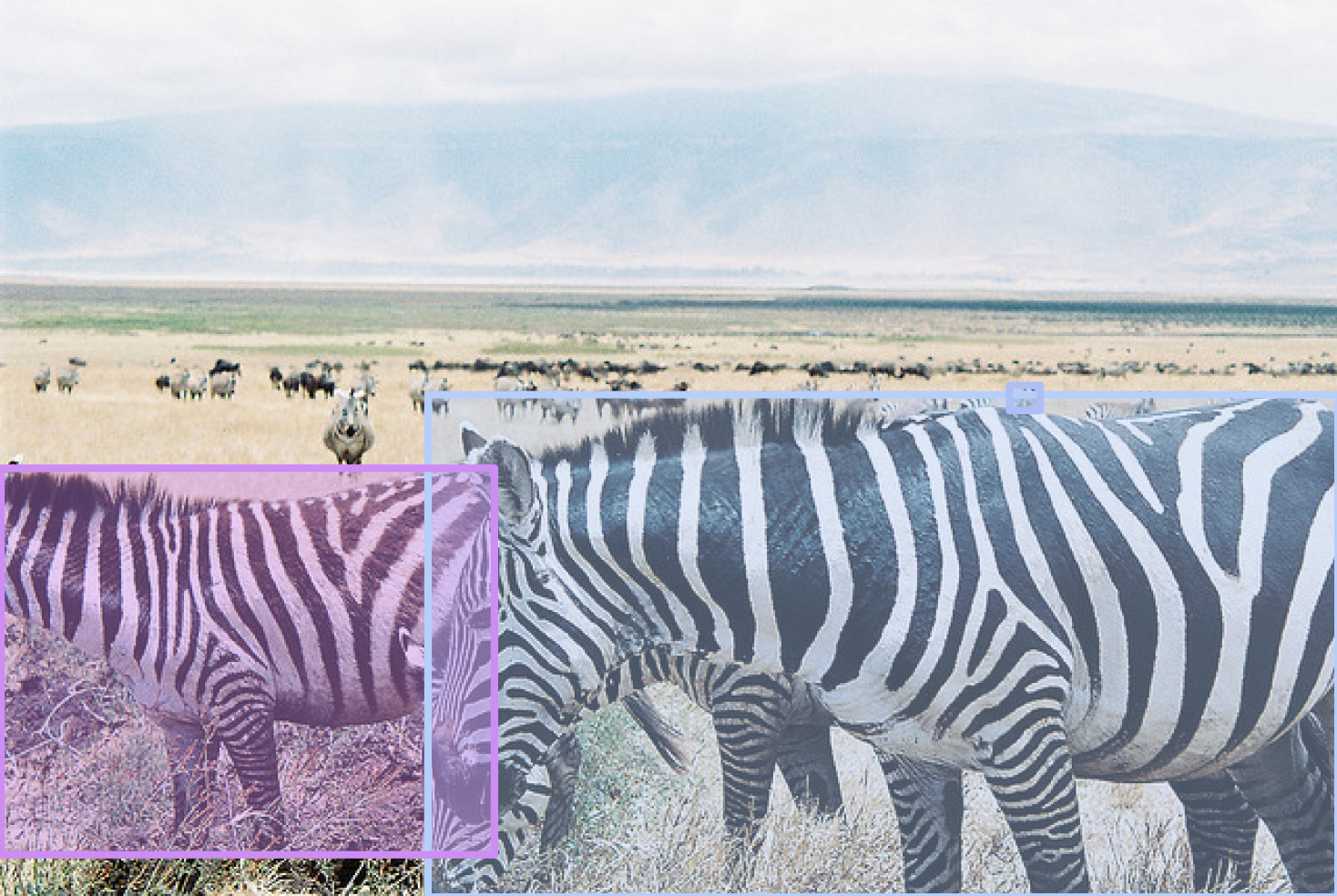}}\hfill
      \subcaptionbox{ID: 22149, toy labeled}{\includegraphics[width=0.495\columnwidth, height=3.5cm]{fig/22149.jpg}}\hfill
      
      \vspace{15pt}
      
      \subcaptionbox{ID: 32206, wrong bounding box}{\includegraphics[width=0.495\columnwidth, height=3.5cm]{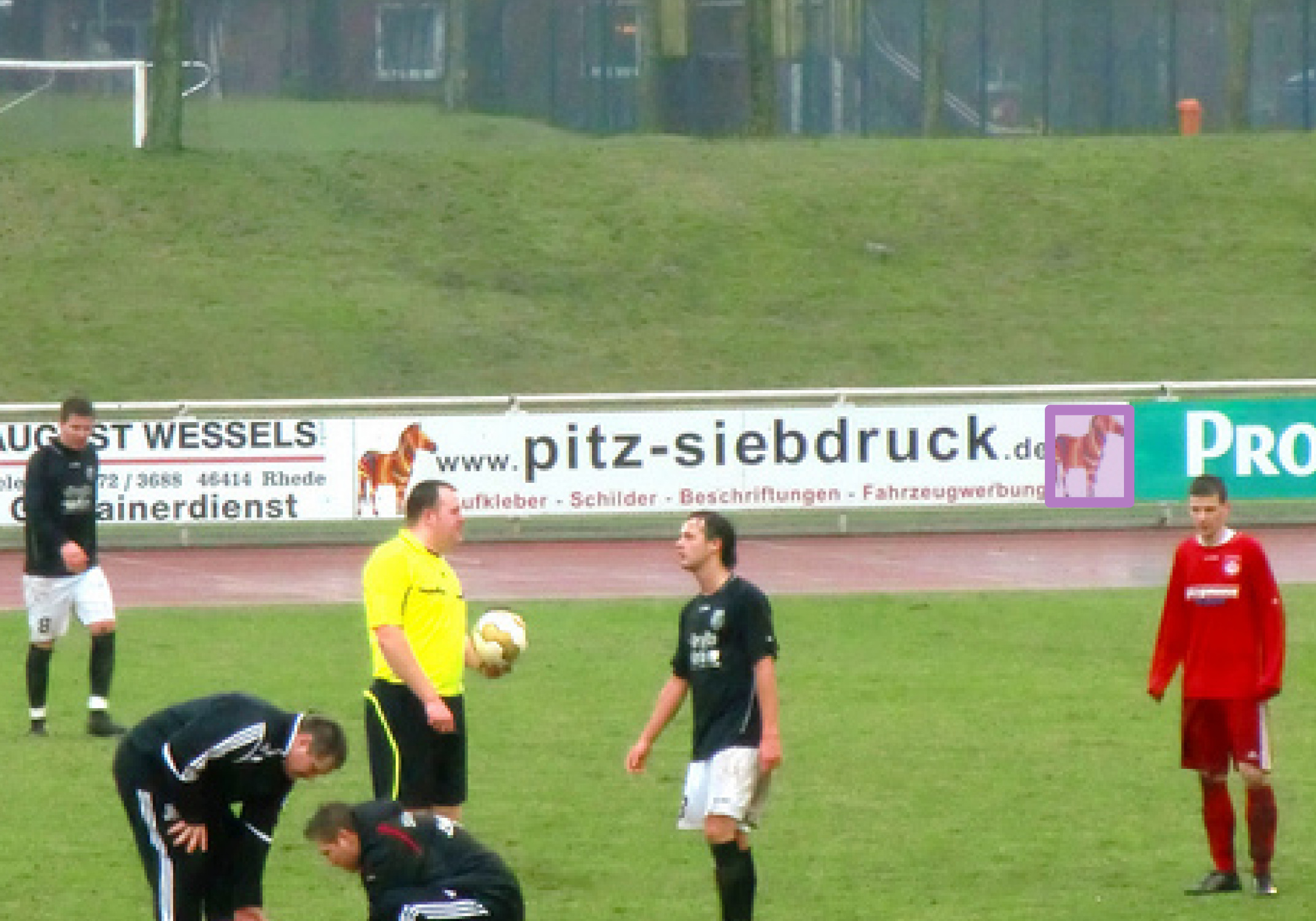}}\hfill
      \subcaptionbox{ID: 533961, imprecise bounding box}{\includegraphics[width=0.495\columnwidth, height=3.5cm]{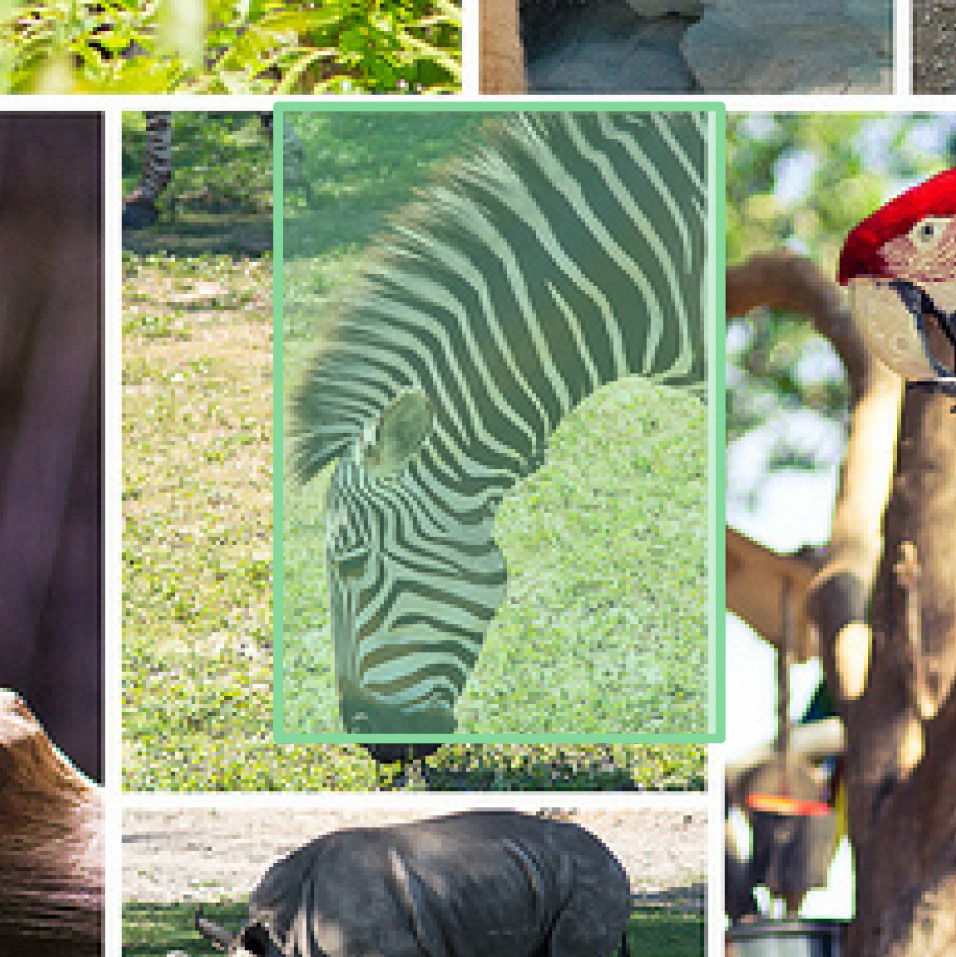}}\hfill
      \vspace{5pt}
    \caption{Four examples of wrongly labelled zebras from the COCO~\cite{cocodataset} dataset. }
    \label{fig:bad-coco}
\end{figure}

\textbf{Animals.} There are not many animals-based datasets available in the literature~\cite{li20214dcomplete, deeplabcut, animalpose}, especially considering full 3D-vertices information and precise segmentation. This is clearly related to the difficulties of collecting and labelling ground truth data in outdoor scenarios. Various approaches have been applied to overcome this problem, ranging from using toy models~\cite{SMAL,SMALR}, merging different datasets~\cite{deeplabcut}, or using synthetic data~\cite{li20214dcomplete,li2021synthetic}. However, all of them fall short in some aspects like lack of animal species variability, size, pose and shape information, skeletal joints location, or limited capturing settings.  For example, \mbox{Horse-10}~\cite{horse10} has only horses moving left-to-right. The authors of the SMAL model~\cite{SMAL} do not release the generated data. The Grévy's zebra dataset~\cite{dpk-zebrads} consists only of 900 low-resolution images that do not contain either correct bounding boxes or labels for all animals. An example of that is provided in Fig.~\ref{fig:bad-dpk}. AnimalPose~\cite{animalpose} focuses on a limited set of animals, in which zebras are not included. The 4DComplete dataset~\cite{li20214dcomplete}, although it contains various animal animations, it fails on releasing textures or textured FBX files making it impossible to customize. They do provide rendered RGB+D images and scene flow but, still, the renderings are provided without any background information. Other synthetic datasets, such as the one from Mu et al.~\cite{Mu_2020_CVPR}, contain data which cannot be used to train a successful detector since they are generated with unrealistic backgrounds and textures~\cite{Mu_2020_CVPR}. These also suffer from the low viewpoint variability and diversity of scenarios, such as the data from COCO. We must also note that, as shown in Fig.~\ref{fig:bad-coco}, COCO is not exempt from wrong or imprecise labelled data.

\textbf{Simulation engines.} Gazebo~\cite{gazebo} is currently the standard for robotic simulation. High reliable physics and tight integration with ROS are its key advantages. However, the lack of visual realism and customization possibility, makes it unusable for generating visual data to be used in learning tasks. Indeed, alternatives emerged in the last years, such as~\cite{airsim, ai2thor, igibson, habitat19iccv, Mller2018, pmlr-v78-dosovitskiy17a}, along with several datasets~\cite{airpose,peoplesanspeople} that use Unreal Engine, Unity, and Blender for rendering. The combination of AirSim and Unreal Engine has been widely explored to generate multiple datasets focused on specific tasks such as human pose estimation~\cite{airpose} and visual odometry~\cite{tartanvo}. Simulators focused on robotics are usually limited by the type of the environment, e.g. indoor~\cite{habitat19iccv,ai2thor}, or the task, e.g. self-driving car~\cite{pmlr-v78-dosovitskiy17a}. Clearly, generalizing them to outdoor scenarios with animated animals is not trivial. GRADE~\cite{GRADE} is a recently introduced method to generate synthetic data built directly upon Isaac Sim. It is a framework that includes both data generation and general robotic testing capabilities thanks to its integration with ROS and the use of ray- and path-tracing. In this work, we leverage the flexibility of GRADE to generate new synthetic data of outdoor scenarios with randomly placed zebras.
	\section{Approach}
\label{sec:approach}
\label{sec:gen}

Using the system introduced in our previous work GRADE~\cite{GRADE}, we generate an outdoor-environment dataset focused on zebras. GRADE is our synthetic data generation framework based on Isaac Sim. Thanks to the flexibility of GRADE, this approach will be easily applicable also to other animal species or setups. The details about the GRADE framework and the simulation management are thoroughly described in~\cite{GRADE}. We proceed here highlighting any major difference with respect to the already introduced system. 

\subsection{Synthetic data}
\subsubsection{Environments}
\label{sec:envdatagen}
We selected nine commercial and one freely available environments from the Unreal Engine marketplace. We used the Unreal Engine Omniverse connector\footnote{\url{https://bit.ly/3X82sph}} to convert them to the USD file format. We list the environments with the corresponding shortened URL in Tab.~\ref{tab:envs}. For each environment, we used directly the available demos and pre-built scenarios. Then, we proceeded to remove the original sky sphere and fix textures when necessary. The connector indeed does not yet support full export of the terrains from Unreal Engine, resulting in a lower level of detail, e.g. missing 3D grass, some textures, and level of details. We replaced the textures with some taken from IsaacSim itself, resembling the \textit{color} of the grass.

\begin{table}[!h]
\centering
\resizebox{\columnwidth}{!}{\begin{tabular}{l|l}
Environment Name & URL                    \\ \hline 
Bliss            & \url{https://bit.ly/3HD3zYP} \\ 
Forest           & \url{https://bit.ly/3mYQv8Z} \\ 
Grasslands       & \url{https://bit.ly/3HD3zYP} \\ 
Iceland          & \url{https://bit.ly/3Ax8zKi} \\ 
L\_Terrain       & \url{https://bit.ly/3V6H7MU} \\ 
Meadow           & \url{https://bit.ly/3Hgxk1n} \\ 
Moorlands        & \url{https://bit.ly/3oHT1ku} \\ 
Rural Australia (Free)  & \url{https://bit.ly/3i5j6Hi} \\ 
Windmills        & \url{https://bit.ly/3AvVTDK} \\ 
Woodland         & \url{https://bit.ly/3mYQv8Z} \\ 
\end{tabular}}
\caption{Names and shortened URLs of the used environments.}
\label{tab:envs}
\end{table}

\subsubsection{Dynamic assets}
\label{subsec:dynamicassets}
We use a freely available zebra model from SketchFab~\cite{sketchfabZebra}. This model consists of 34 different in-place animation sequences, i.e. without root translation or rotation movements, for a total of 888 animation frames. We converted each animation sequence to the USD format using Blender and its Omniverse connector. Then, we post-processed the sequence to obtain per-frame vertices position and skeletal information. This allows us, for every generated frame, to have corresponding ground truth information about these two characteristics. The vertices are used to compute oriented bounding boxes that are then employed for the placement procedure.

\subsubsection{Placement of zebras}
Zebra placement is based on an ad-hoc procedure that is repeated every time a frame is generated. For every environment we select a specific mesh as `terrain', which represents the area in which we will then place the zebras. The placement consists then of four main steps: i) selecting a random rectangular area of the terrain, ii) randomly selecting a set of zebras, iii) for each zebra select frame of its animation sequence, a scaling factor and a global orientation of the zebra, iv) place the zebra in the rectangular area considering the bounding-box occupancy. The sides of the rectangle are randomly selected to be between 40 and 120 meters, while the scaling factor ranges between 40\% and 100\% and allows us to obtain a higher degree of variability. The placement is an iterative procedure that considers one zebra model after the other. Any model that cannot be placed following a detected collision is removed from the simulation. The final results depend mostly on the resolution of the terrain mesh for both collisions between meshes and contact of the zebras with the ground. In general, we noted that collisions are rare and that contacts with the ground are good. Note that, as opposed to~\cite{GRADE}, we do not consider the full animation sequence since we lack any root translation information.

\subsubsection{Data collection methodologies}
\label{sec:coll}
Contrary to what is done for indoor environments in~\cite{GRADE}, here we focus on image generation rather than video sequences. We also perform a series of randomization for each captured frame, i.e. the i) time of day, ii) number of zebras in the environment, iii) their scaling factor, iv) their specific animation frame, and the v) placement of three cameras that will record the scene. Specifically, given any environment, we set up three aerial cameras and randomly pre-load 250 zebras at the beginning of each experiment. We then uniformly select the number of zebras that will be placed in the next frame. This number is set to be between 2 and 250. Note that this is \textit{not} the number that will appear in each frame, nor is the final number of zebras that are actually placed. As explained above, the placement strategy may remove some of the zebras and the camera may not observe all the zebras given a point of view. Once the placement happened, we randomize the location of the cameras and the time of days three times. The time of day will be 90\% of the time between 5 am and 8 pm, which results in good lighting conditions given our current settings, and 10\% of the time in the remaining hours, resulting in dusk-to-night light settings. This further randomizes the appearances of both the generated frames and the shadows. Cameras are placed using the average location of the zebras as a pivot point. For the placement of the cameras, we distinguish between two slightly different image-generation procedures: one more general and one more focused on capturing zebras from a nearer viewpoint. We first describe the former and then identify the minor modifications that we applied to the latter. From the pivot point, we randomize the distance in the x-y plane and the height of the camera. The height is set to be between 5 and 20 meters more than the average of the zebras, while the x and y are set to be between -/+ 100 meters. Once the position is fixed, we can compute and randomize roll, pitch, and yaw. Roll is set to be within $[-10,10]$ degrees, yaw is set to be the ray that connects the camera and the pivot point with an additional random $[-30,30]$ degrees. Pitch is computed as $\theta = atan2(pivot_z - cam_z, d(pivot, camera)) + 15$ degrees, where $d(pivot, camera)$ is the distance in the x, y plane. The modifications applied to this methodology during the second image-generation procedure are as follows: the x and y positions are set to be within 5 meters of the virtual bounding box containing all zebras, the yaw has an additional $[-15,15]$ degrees component instead of the $[-30,30]$. This results in images that are closer to the zebras than the ones obtained from the former camera placement strategy.

For each environment, we randomly place the zebras 200 times, resulting in 600 frames per experiment per camera, i.e. 1.8K frames in total. After the generation process is complete for all ten environments, this totals to 18K frames captured with the first camera placement strategy, and 18K with the camera set to be more nearby, totalling 36K frames. For each frame, we save the pose of the cameras, zebra skeletal pose and meshes vertices, ground truth depth and instance segmentation.

\subsection{Real-world data}
\label{sec:realdata}
\begin{figure*}[!ht]
  
  \subcaptionbox{R1}{\includegraphics[width=0.495\textwidth, height=5cm]{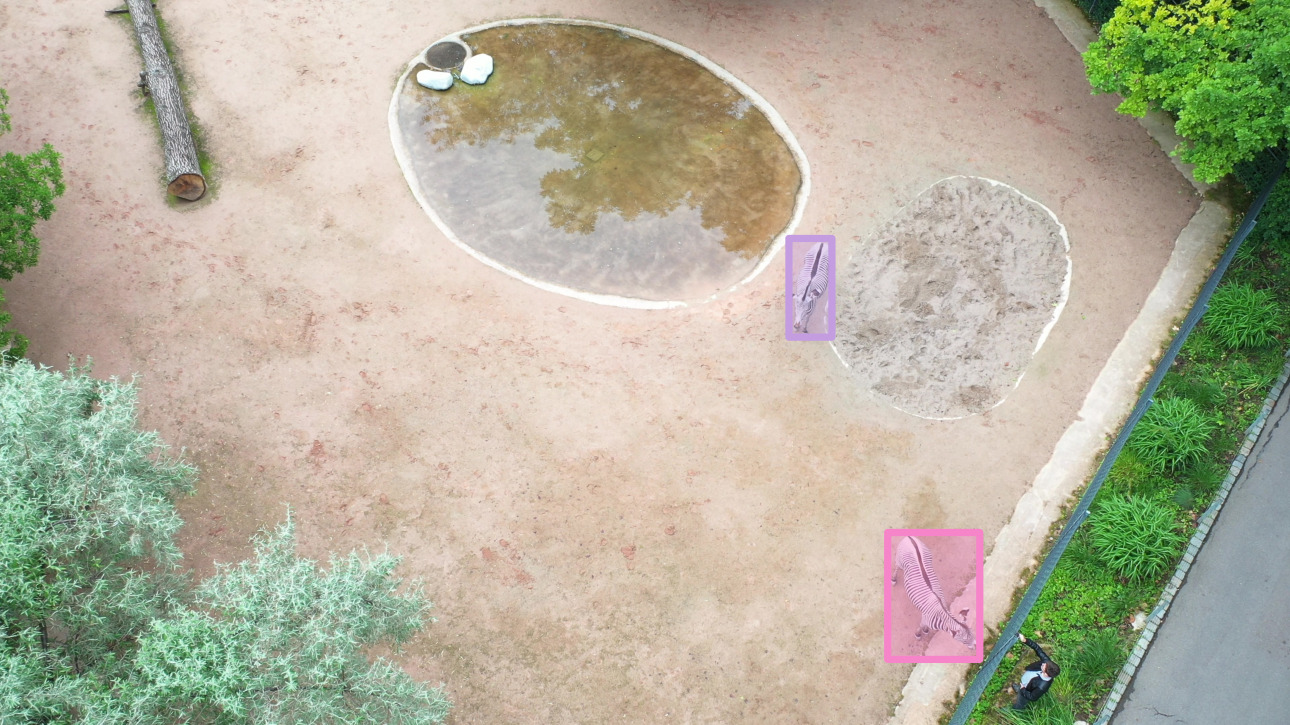}}\hfill
  \subcaptionbox{R2}{\includegraphics[width=0.495\textwidth, height=5cm]{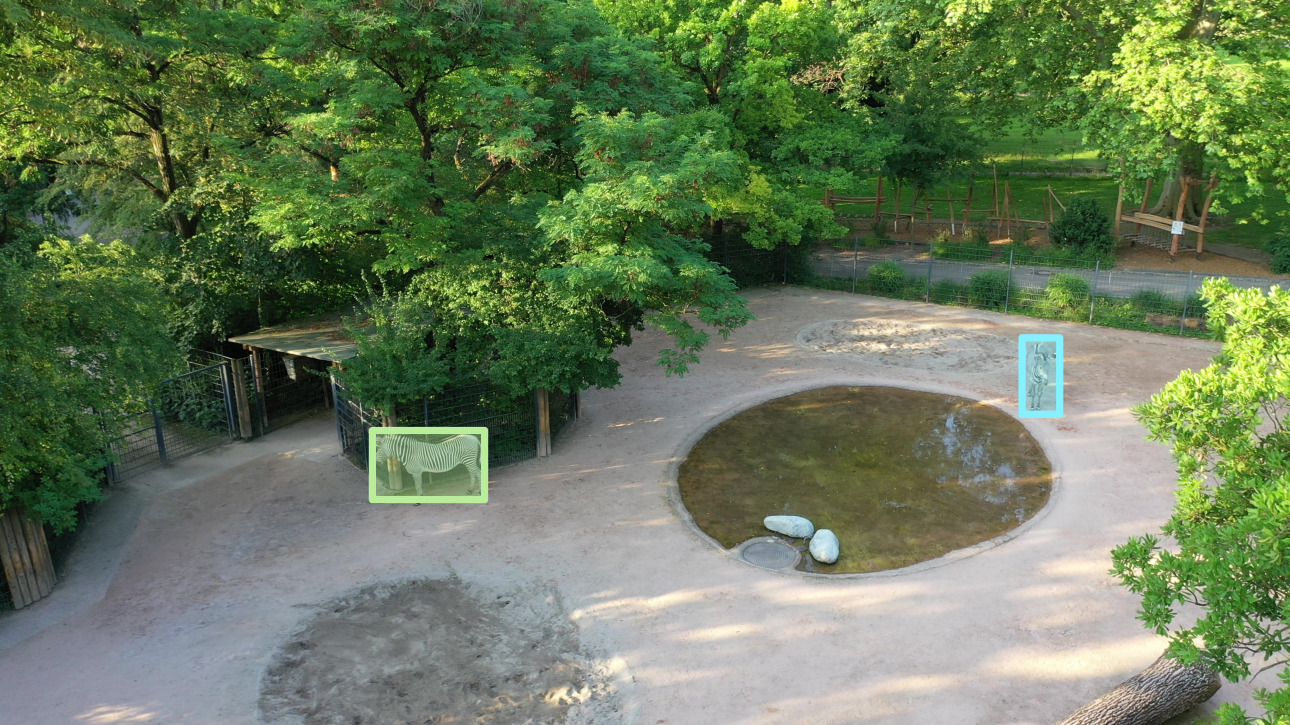}}\hfill \\
  
  \vspace{10pt}
  
  \subcaptionbox{R3}{\includegraphics[width=0.495\textwidth, height=5cm]{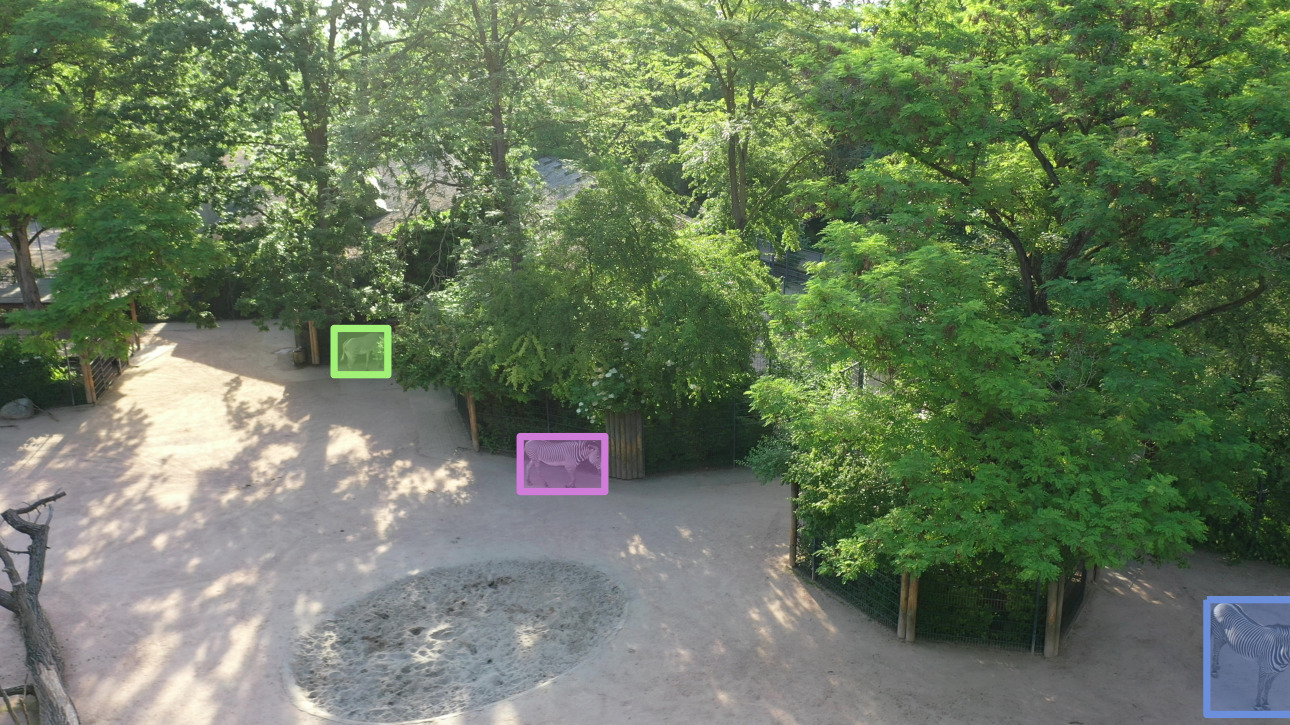}}\hfill
  \subcaptionbox{RP}{\includegraphics[width=0.495\textwidth, height=5cm]{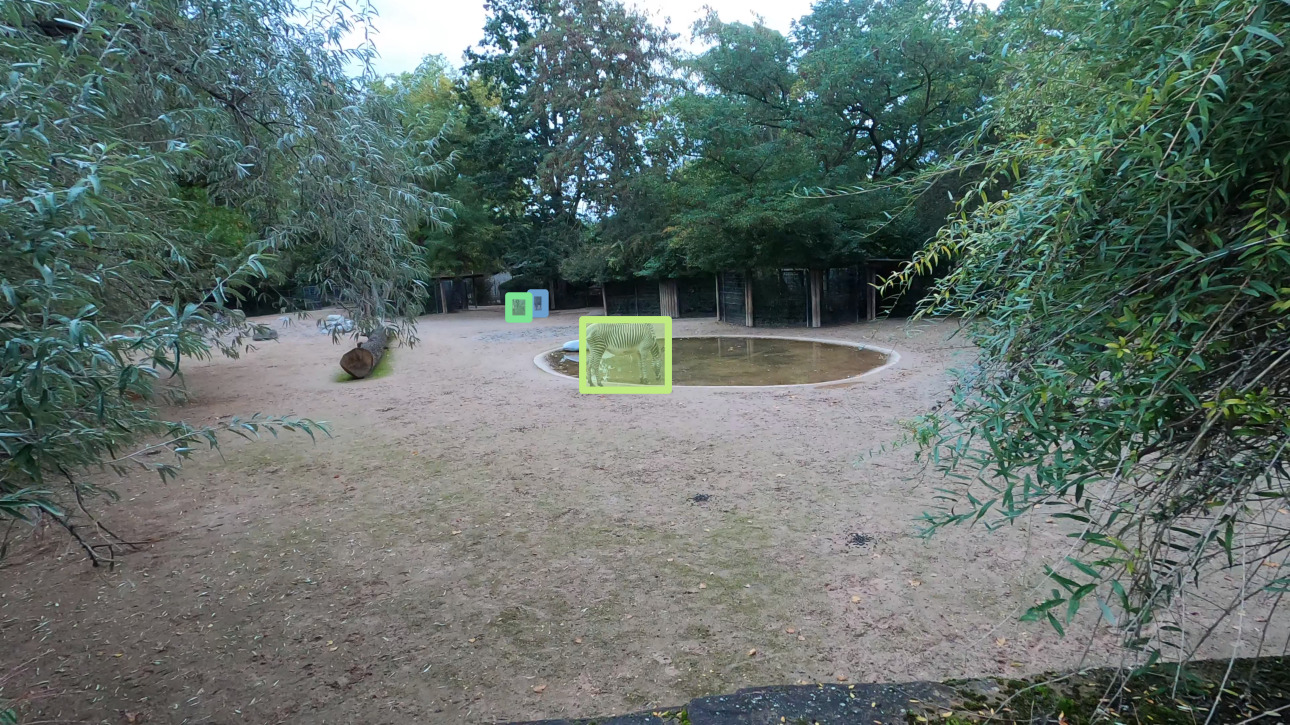}}\hfill
  
  \vspace{10pt}
  
  \caption{An example of our real-world collected images used for testing. Three aerial and one `ground-level' views. Best viewed in color.}
  \label{fig:labels-wil}
\end{figure*}

We performed several data collection experiments in a controlled scenario within the Wilhlema Zoo in Stuttgart (DE). An example of the collected data can be found in Fig.~\ref{fig:labels-wil}. We used two manually flown DJI Mavic 2 Pro drones, recording images at 29.97 fps at a resolution of $3840\times2160$, and three GoPro Hero8, also with a resolution of $3840\times2160$ at 59.94 fps. None of the GoPros had fixed locations between experiments. The data has been manually synchronized by using a recorded light signal visible by all cameras at the same time. We then extracted one frame every five seconds from all the videos. Out of these, 905 images were randomly selected and annotated manually and precisely. These annotations were then used to train an SSD multibox~\cite{ssd} detector, which, with \textit{Smarter-Labelme}~\cite{price2023accelerated}, allowed us to obtain bounding boxes on our video sequences with ease. Out of all the data available, we finally selected three collections during which the zebras were visible by both drones. The boxes on those sequences were then manually refined in a final step. This procedure thus resulted in 905 precisely annotated images, and 104K frames annotated with~\cite{price2023accelerated}. Within this work, we release the data used during our training experiments, i.e. the 905 precisely annotated images and 200 automatically labelled ones from one of the experiments (see Sec.~\ref{sec:exp} for details).

	\section{EXPERIMENT AND EVALUATIONS}
\label{sec:exp}

Here we seek to demonstrate that the synthetic data generated by our method can be used effectively for a vision-based task which is highly related to image features and context, such as the detection of zebras in outdoor wild environments from an aerial point of view. Our hypothesis is that, by training a model using only synthetic data acquired in a realistic simulation environment, we can achieve detection performance on real images comparable to a model trained on a manually and very-precisely labelled set of real images. Our goal is to prove that synthetically generated data \textit{alone} can be used to train a network capable of detecting zebras with high accuracy in real-world images. To that end, we decided to perform various tests on YOLOv5s. We train the networks from scratch and with mixed datasets to test the performance and provide a complete overview. All the training runs are made from scratch with the default hyperparameters and for the standard 300 epochs. We do not introduce any additional data augmentation technique different from the one applied by default by the YOLOv5 code. This consists of some randomization in the scale, horizontal flip, translation and HSV colour space factors. We do not modify these values to have a fairer comparison across the models that would not require parameter grid searches or other steps when compared to the baseline pre-trained model. We save the best model, as evaluated on the specific validation set, and compare it over multiple datasets. We evaluate the performance with the COCO standard metric (mAP@[.5, .95], AP in this work) and the PASCAL VOC's metric (mAP@.5, AP50 in this work). We also report the average and weighted average of these two metrics. We weigh based on the cardinality of each one of the evaluated datasets. With these comparisons, we demonstrate that, with our synthetic data, we can successfully capture real-world features. This, while also obtaining trained models which show, in general, improved performance when compared to the pre-trained ones.

\begin{figure}[!h]
  \includegraphics[width=0.24\textwidth, height=2.8cm]{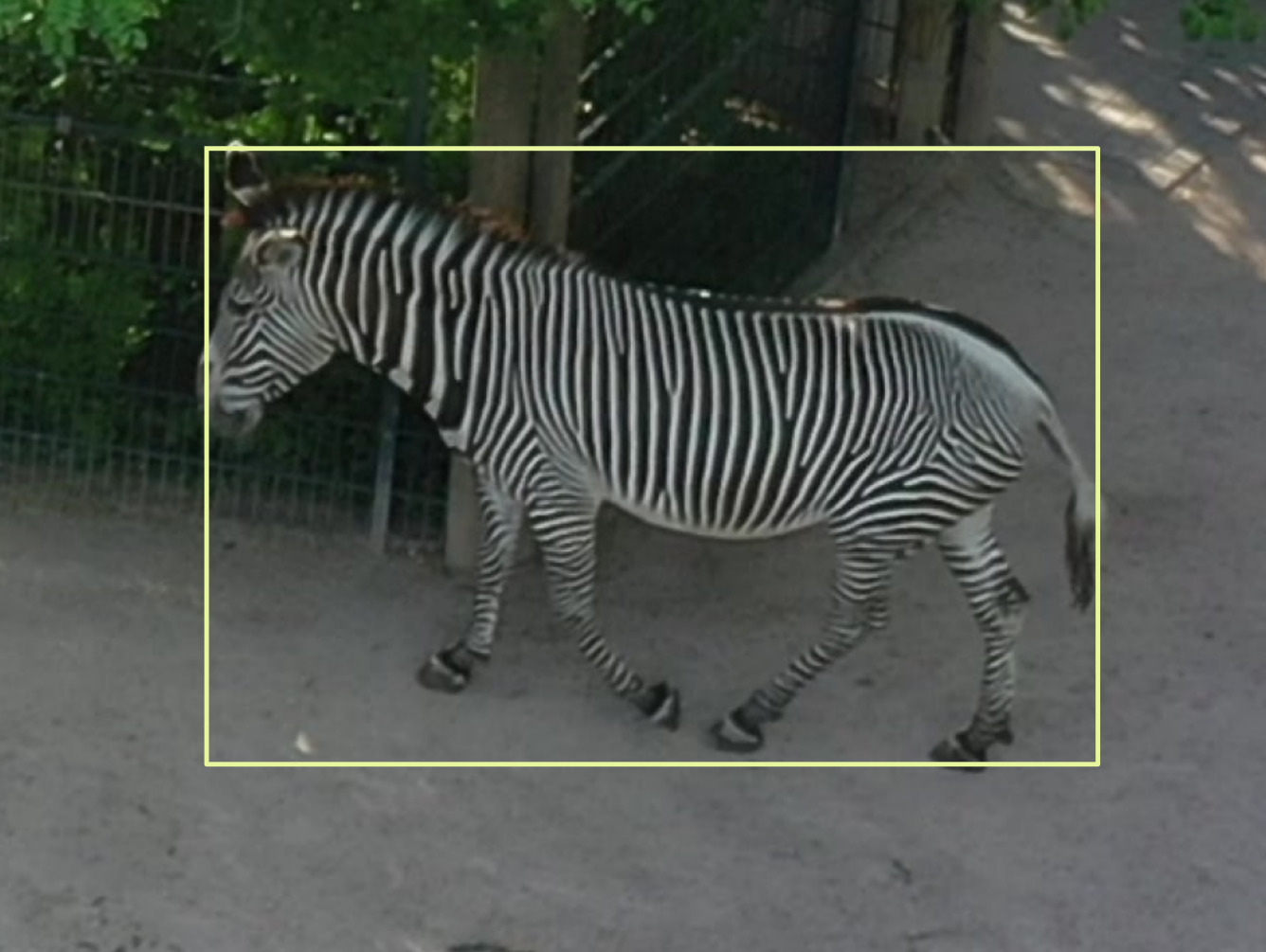}
  \includegraphics[width=0.24\textwidth, height=2.8cm]{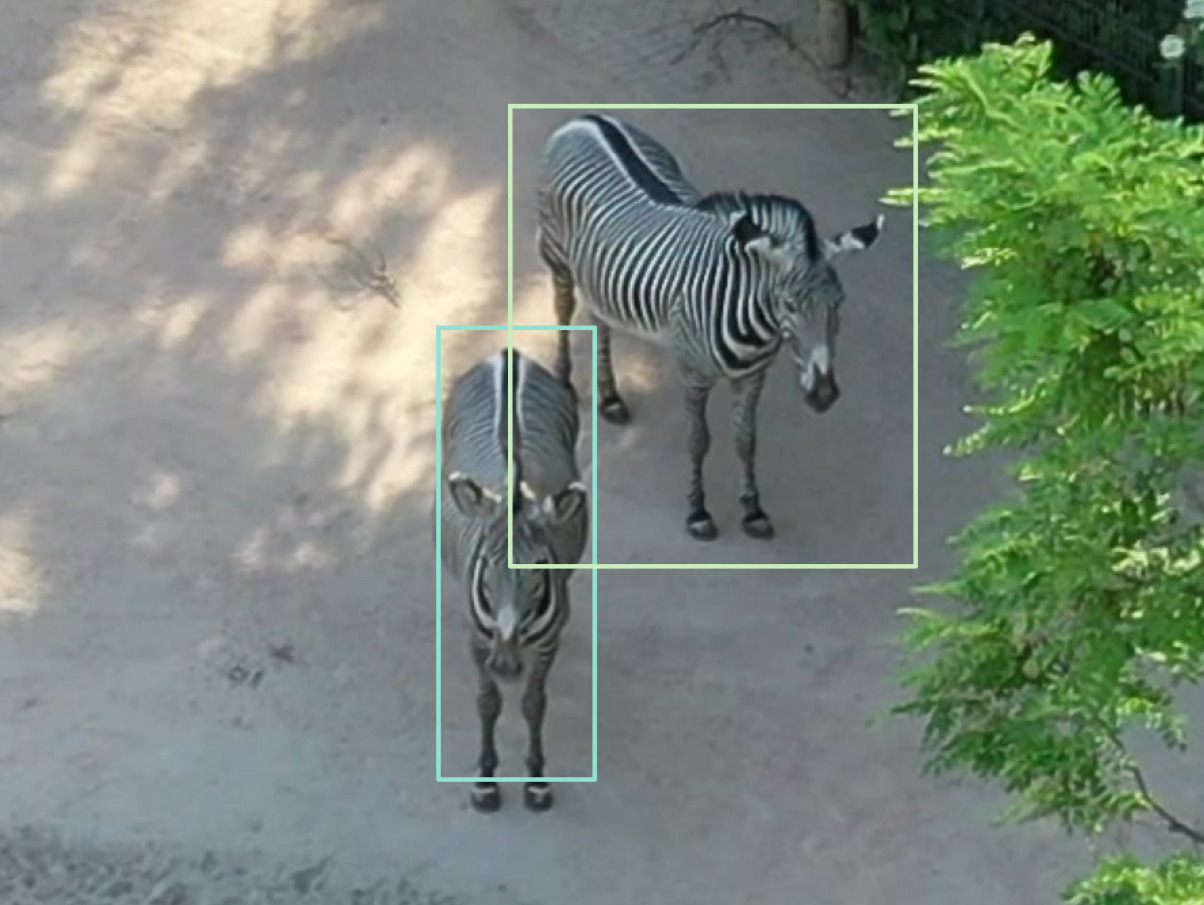}
  \caption{Two zoomed-in examples of imprecisely labelled data. The bounding boxes can either be slightly too loose or too tight on the zebras.}
  \label{fig:imprecise}
\end{figure}

Synthetic Full dataset (SF) is the dataset containing all the 36K synthetically generated images. These are then randomly shuffled and split into 80/20 train/validation sets. Synthetic Closeby (SC) is the synthetic data generated only by the second strategy, as described in Sec.~\ref{sec:coll}, i.e. 18K images for which the camera is within 5 meters of the bounding box containing all the zebras. This data is also divided randomly with an 80/20 ratio. With COCO we refer to the images of the COCO dataset~\cite{cocodataset} which contains zebras, i.e. 1916 training and 85 validation examples. Due to the small size of the validation set of the COCO dataset, we do not perform any training on this data alone. With APT-36K we indicate the set of images from the APT-36K~\cite{apt-36k} dataset which contains zebras, i.e. 1.2K samples. We then have, R1, R2, and R3 which are three sets of real-world data which is not precisely labelled, as described in Sec.~\ref{sec:realdata}. To distinguish between which drone captured the given sequence, we use the suffixes \textit{\_D1} and \textit{\_D2}. R1 consists of 19.7K images, R2 of 23.4K, and R3 of 8.8K, for each drone. Finally, we use RP to indicate the set of the 905 real-world images precisely labelled by us, sampled from representative images from the previous Rx datasets and additional images captured with the GoPros as described in Sec.~\ref{sec:realdata}. Of them, 720 are randomly used in training and 185 for validation. An example of the bounding boxes of our real-world data is provided in Fig.~\ref{fig:labels-wil}. We also provide two zoomed-in examples of imprecise labels in Fig.~\ref{fig:imprecise}. Note that also other datasets, e.g. COCO (see Fig.~\ref{fig:bad-coco}), present such approximations.

\begin{table}[h]
    \centering
    \resizebox{\columnwidth}{!}{
\begin{tabular}{l|c|c|c}
  Dataset & Description & Train imgs. & Val imgs. \\ \hline
 SF & Our full synthetic data & 29K & 7K \\
 SC & \begin{tabular}{c}A subset of our synthetic data \\ focused on camera poses closer to the zebras\end{tabular} & 14.5K & 3.5K \\ 
 R1 & Aerial capture experiment 1 & --- & 19.7K \\
 R2 & Aerial capture experiment 2 & --- & 23.4K \\
 R3 & Aerial capture experiment 3 & --- & 8.8K \\
 R3$_\text{100}$ & 100$\times$2 images randomly chosen from R3 & 100 & 100 \\ 
 RP & Preciselly labelled images from R1/2/3 and GoPros  & 720 & 185 \\
 COCO & COCO-zebras & 1916 & 85 \\ \hline
 Rx\_D1,2 & \multicolumn{3}{c}{Either 1st or 2nd drone capturing during experiment Rx} \\
 $\ast$-1920 & \multicolumn{3}{c}{Same dataset but using 1920 image size during training}\\
 $\ast$ + $\diamond$ & \multicolumn{3}{c}{Trained by merging $\ast$ and $\diamond$ corresponding train and validation sets} \\
\end{tabular}}
\caption{Zebras datasets legend and training/validation sizes}
    \label{tab:dataset-size}
\end{table}

\begin{table*}[!ht]
\resizebox{\textwidth}{!}{%
\begin{tabular}{l|cc|cc|cc|cc|cc|cc|cc|cc|cc|cc}
 & \multicolumn{2}{c|}{APT-36K~\cite{apt-36k}} & \multicolumn{2}{c|}{R1\_D1} & \multicolumn{2}{c|}{R1\_D2} & \multicolumn{2}{c|}{R2\_D1} & \multicolumn{2}{c|}{R2\_D2} & \multicolumn{2}{c|}{R3\_D1} & \multicolumn{2}{c|}{R3\_D2} & \multicolumn{2}{c|}{RP (validation)} & \multicolumn{2}{c|}{Weigthed avg.}  & \multicolumn{2}{c}{Avg.} \\ \cline{2-21}
Training Dataset & mAP50 & mAP & mAP50 & mAP & mAP50 & mAP & mAP50 & mAP & mAP50 & mAP & mAP50 & mAP & mAP50 & mAP & mAP50 & mAP & mAP50 & mAP & mAP50 & mAP \\ \hline
SF & 0.072 & 0.029 & 0.770 & 0.488 & 0.756 & 0.490 & 0.224 & 0.130 & 0.597 & 0.393 & 0.142 & 0.092 & 0.203 & 0.127 & 0.104 & 0.074& 0.498 & 0.318 & 0.359 & 0.228\\
SF-1920 & 0.103 & 0.055 & 0.853 & 0.568 & 0.958 & 0.646 & 0.873 & 0.540 & 0.957 & 0.616 & 0.608 & 0.443 & 0.651 & 0.484 & 0.287 & 0.191 & 0.853 & 0.563 & 0.661 & 0.443 \\
SC & 0.121 & 0.046 & 0.714 & 0.455 & 0.830 & 0.529 & 0.147 & 0.084 & 0.513 & 0.316 & 0.156 & 0.097 & 0.375 & 0.202 & 0.092 & 0.061& 0.482 & 0.299 & 0.369 & 0.224 \\
SC-1920 & 0.150 & 0.053 & 0.907 & 0.605 & 0.971 & 0.664 & 0.939 & 0.593 & 0.968 & 0.652 & 0.649 & 0.476 & 0.819 & 0.580 & 0.331	& 0.228 & {0.901} & {0.604} & 0.717 & 0.481 \\
RP & 0.260 & 0.092 & 0.865 & 0.487 & 0.935 & 0.550 & 0.808 & 0.479 & 0.946 & 0.593 & 0.772 & 0.380 & 0.922 & 0.548 & 0.805 & 0.453& 0.873 & 0.512 & 0.789 & 0.448\\ 
RP-1920 & 0.161 & 0.066 & 0.937 & 0.615 & 0.980 & 0.663 & \textbf{0.989} & 0.653 & 0.982 & 0.666 & 0.801 & 0.532 & \textbf{0.986} & 0.680 &\textbf{0.914} & \textbf{0.636} & 0.950 & 0.636 & 0.844 & 0.564 \\ \hline
SC+COCO-1920 & 0.709 & 0.386 & 0.943 & 0.624 & 0.976 & 0.676 & 0.932 & 0.599 & 0.977 & 0.659 & 0.637 & 0.481 & 0.867 & 0.635 & 0.350 & 0.253 & 0.918 & 0.621 & 0.799 & 0.539 \\
RP+COCO-1920 & 0.837 & 0.526 & 0.967 & 0.639 & 0.984 & 0.681 & 0.980 & \textbf{0.656} & 0.968 & 0.684 & 0.768 & 0.493 & 0.981 & 0.674 & 0.911 & 0.626& 0.956 & 0.650  & \textbf{0.925} & \textbf{0.622}\\
SC+COCO+$\text{R3}_{100}$-1920 & 0.704 & 0.378 & \textbf{0.975} & \textbf{0.655} & \textbf{0.994} & \textbf{0.707} & 0.963 & 0.636 &\textbf{0.991} & \textbf{0.691} & \textbf{0.986} & \textbf{0.733} & 0.961 & \textbf{0.708} & \underline{0.432} & \underline{0.308} & \textbf{0.975} & \textbf{0.676}  & 0.878 & 0.602\\ \hline
SC+COCO+RP-1920 & 0.705 & 0.383 & 0.988 & 0.688 & 0.994 & 0.714 & 0.988 & 0.652 & 0.990 & 0.709 & 0.869 & 0.614 & 0.988 & 0.756 & 0.921 & 0.639 & 0.976 & 0.685 & 0.930 & 0.644 \\ \hline
Pretrained-COCO & \textbf{0.879} & \textbf{0.566} & 0.576 & 0.376 & 0.529 & 0.354 & 0.421 & 0.274 & 0.379 & 0.258 & 0.331 & 0.215 & 0.551 & 0.390 & 0.173 & 0.123 & 0.469 & 0.312 & 0.480 & 0.320
\end{tabular}
}
\caption{Results of the evaluations of the trained models. We report mAP50 and mAP for each dataset as well as both the average and weighted average of these metrics. We divide between models trained on mixed datasets, vanilla ones, and the model pre-trained with COCO. In bold the best results. We underline the best model not using the RP dataset during training in the corresponding validation column.}
\label{tab:res}
\end{table*}

Our baseline for comparison consists of the network pre-trained on the full COCO dataset. We perform some training tests on both the default $640\times640$ image size and the increased $1920\times1920$, identified in our table with the `-1920' suffix. Once established that the bigger image size yields better results, we trained the network with mixed datasets. These are i) SC+COCO-1920, which combines SC training and validation sets with images from COCO's corresponding splits, ii) SC+COCO+$\text{R3}_{100}$-1920, which adds 100 train and 100 validation images randomly sampled from R3, and iii) RP+COCO-1920, which merges RP and COCO sets. Note that all models trained with RP have been exposed to representing data coming from Rx\_Dx, giving them an advantage in these evaluations. The full description of the datasets, including the training a validation set sizes, is reported in~\ref{tab:dataset-size}.

\begin{figure*}[!ht]
  
  \includegraphics[width=0.195\textwidth, height=2.8cm]{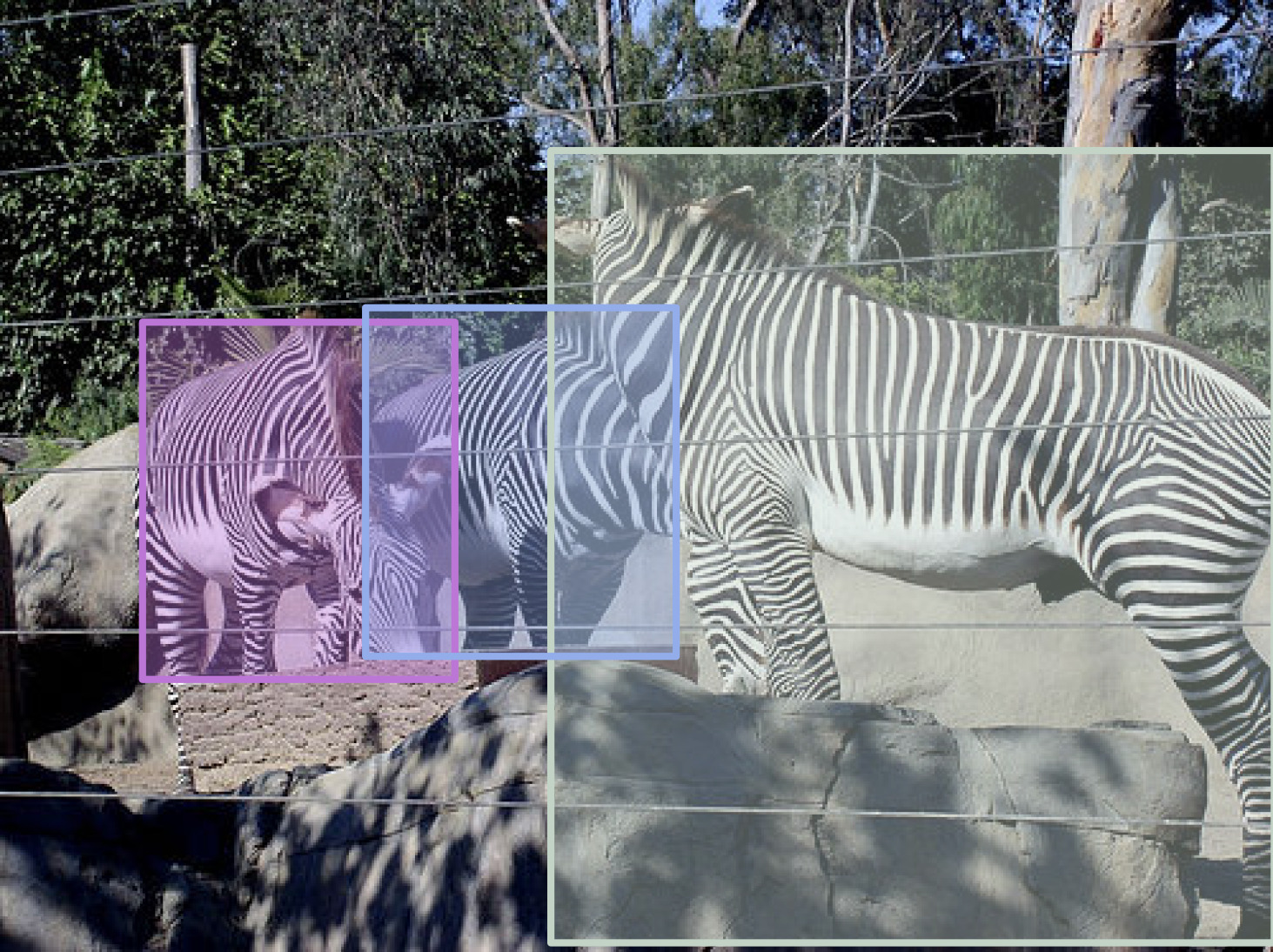}\hfill
  \includegraphics[width=0.195\textwidth, height=2.8cm]{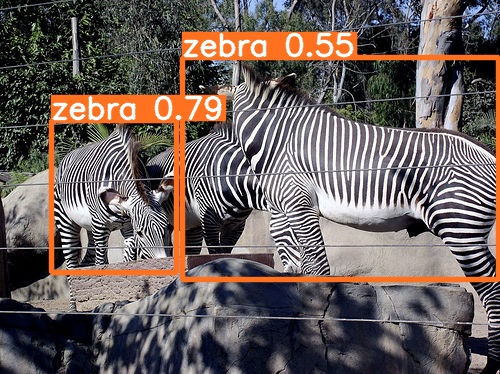}\hfill
  \includegraphics[width=0.195\textwidth, height=2.8cm]{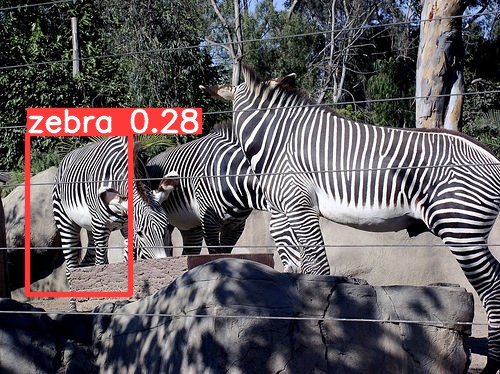}\hfill
  \includegraphics[width=0.195\textwidth, height=2.8cm]{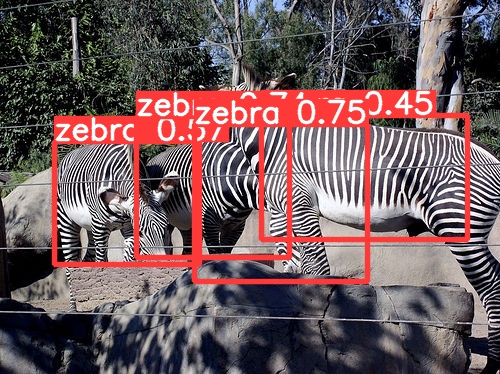}\hfill
  \includegraphics[width=0.195\textwidth, height=2.8cm]{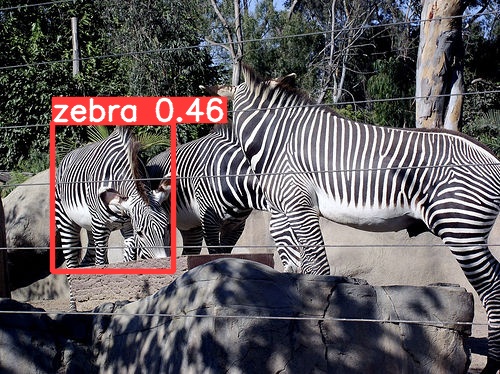}\hfill 

    \vspace{3pt}

  \includegraphics[width=0.195\textwidth, height=2.8cm]{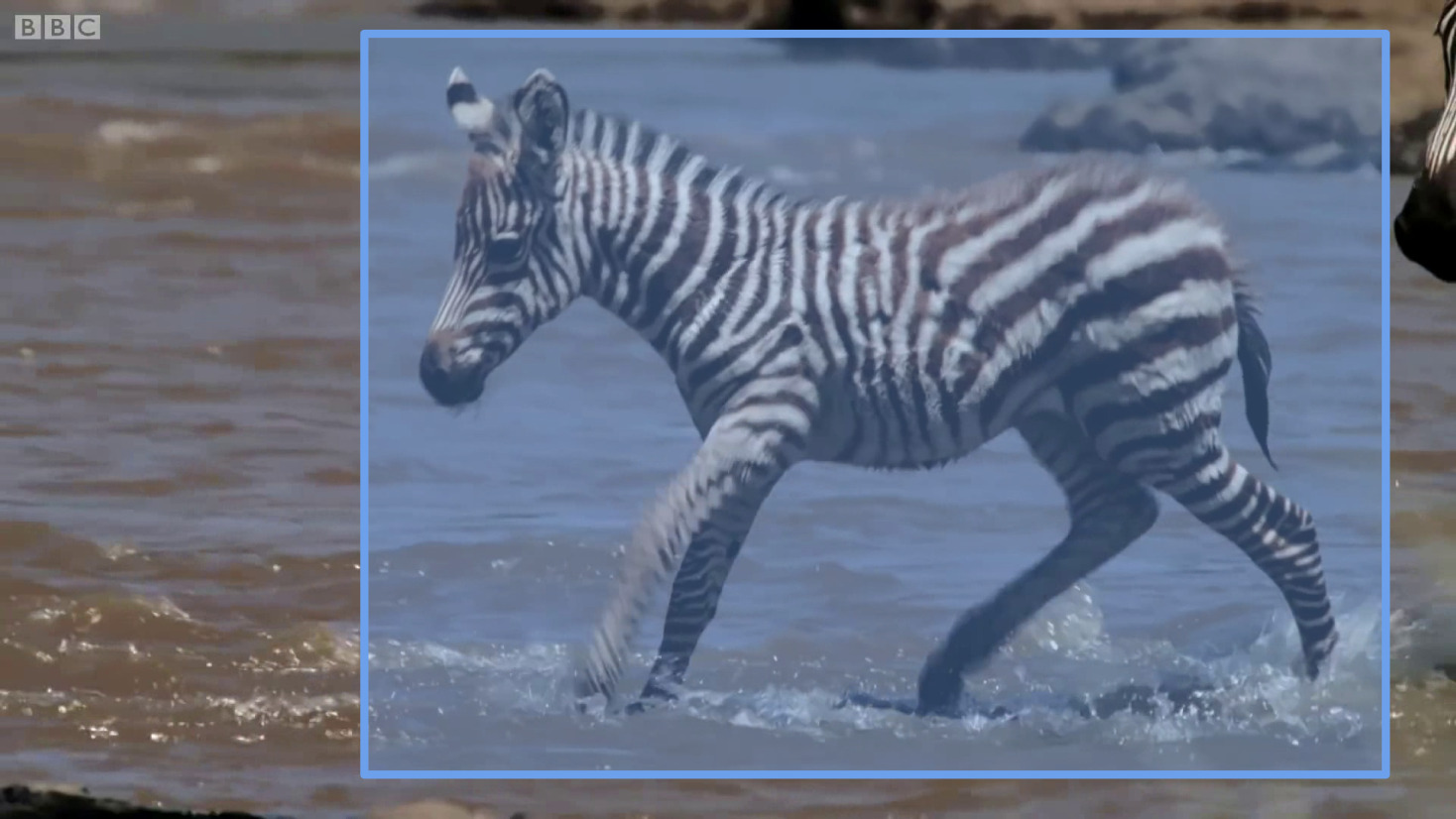}\hfill
  \includegraphics[width=0.195\textwidth, height=2.8cm]{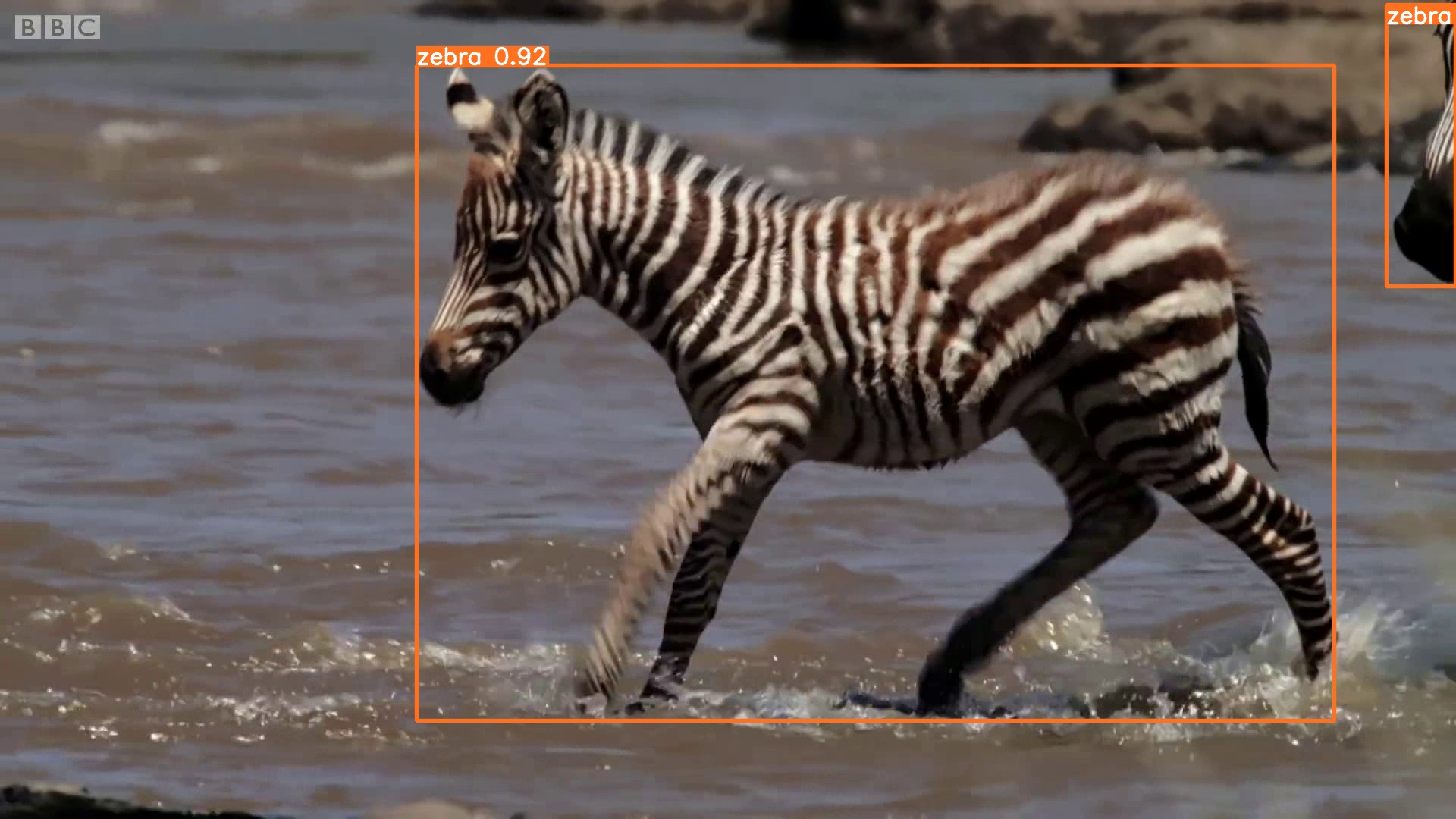}\hfill
  \includegraphics[width=0.195\textwidth, height=2.8cm]{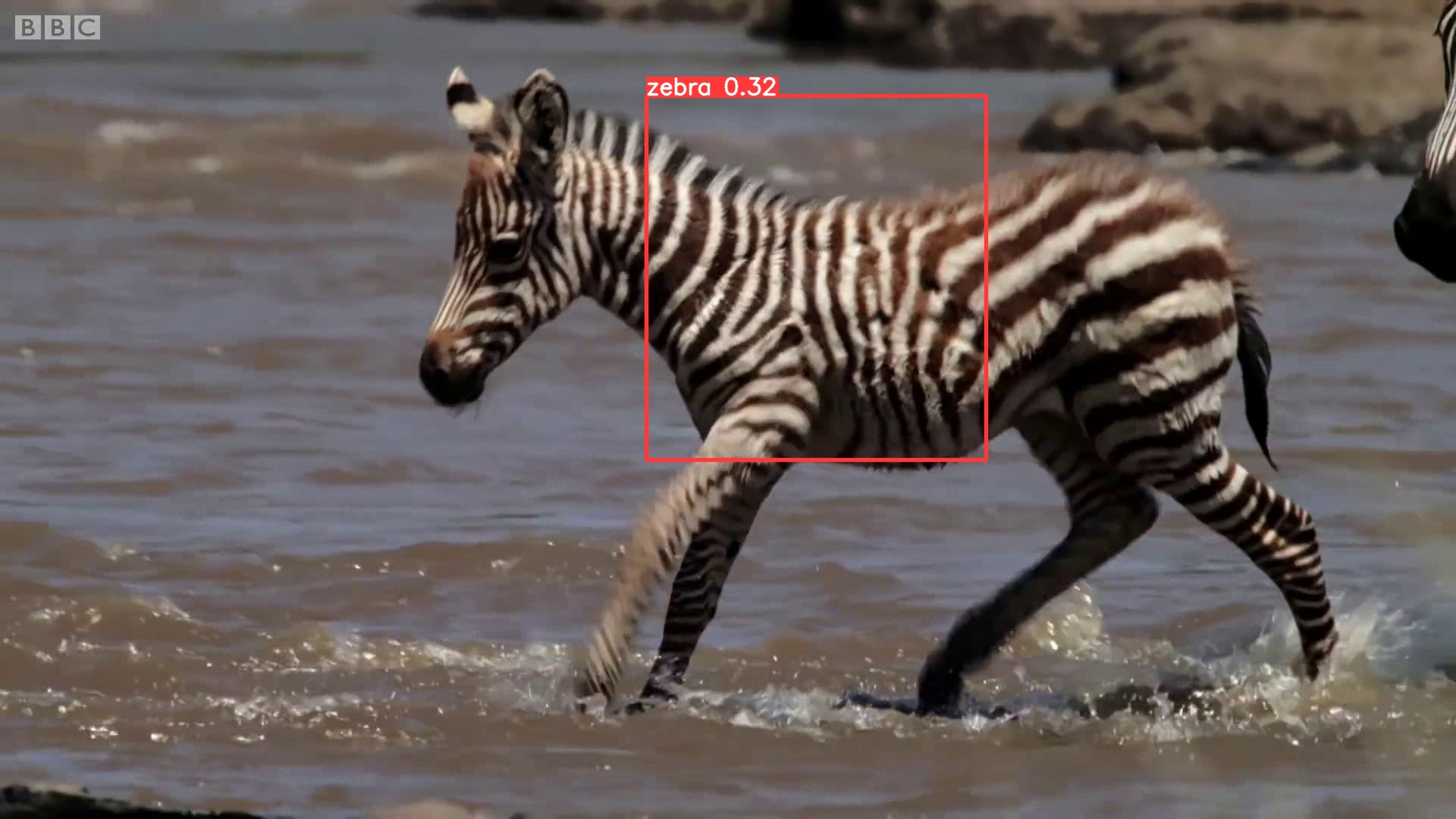}\hfill
  \includegraphics[width=0.195\textwidth, height=2.8cm]{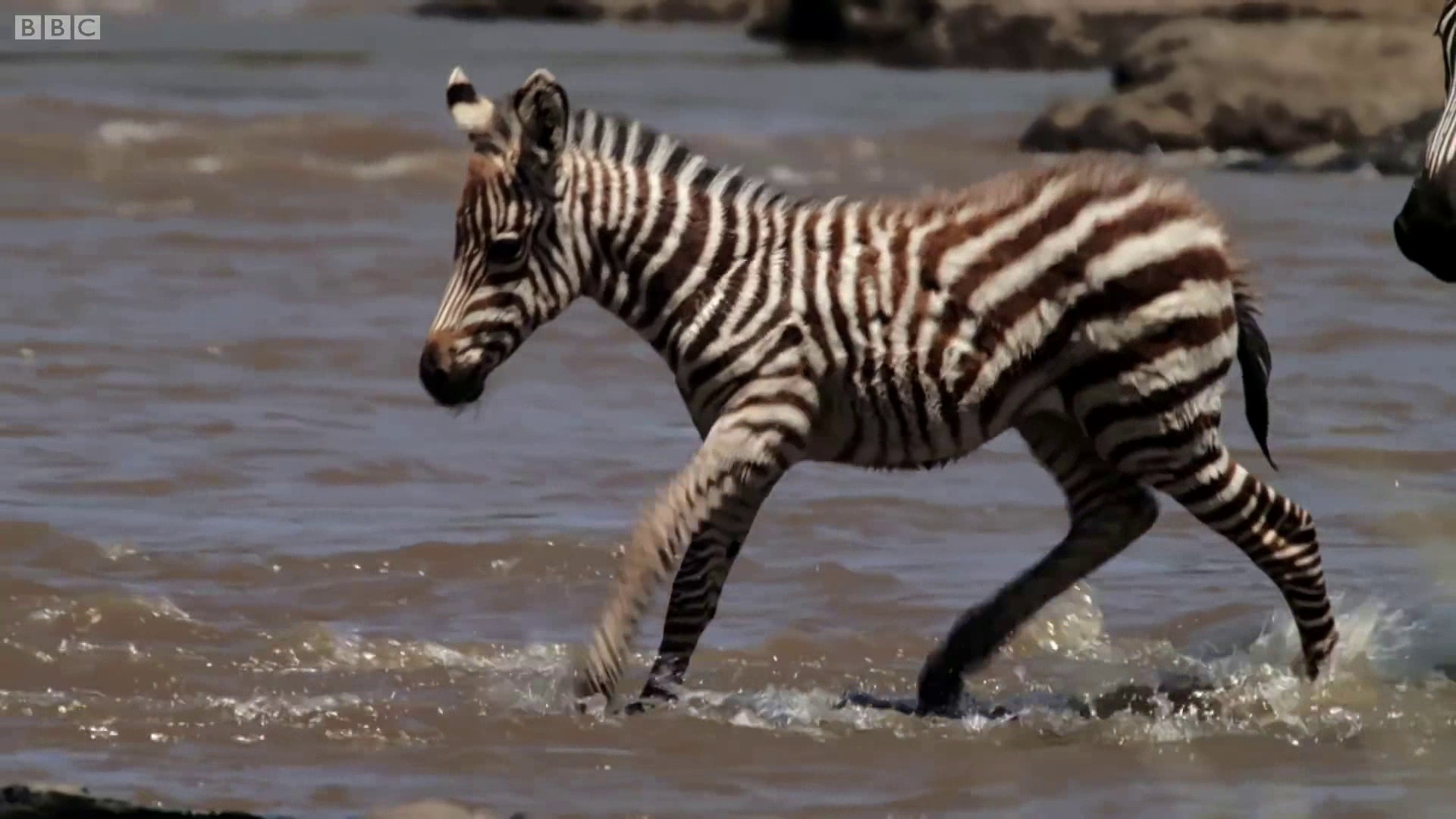}\hfill
  \includegraphics[width=0.195\textwidth, height=2.8cm]{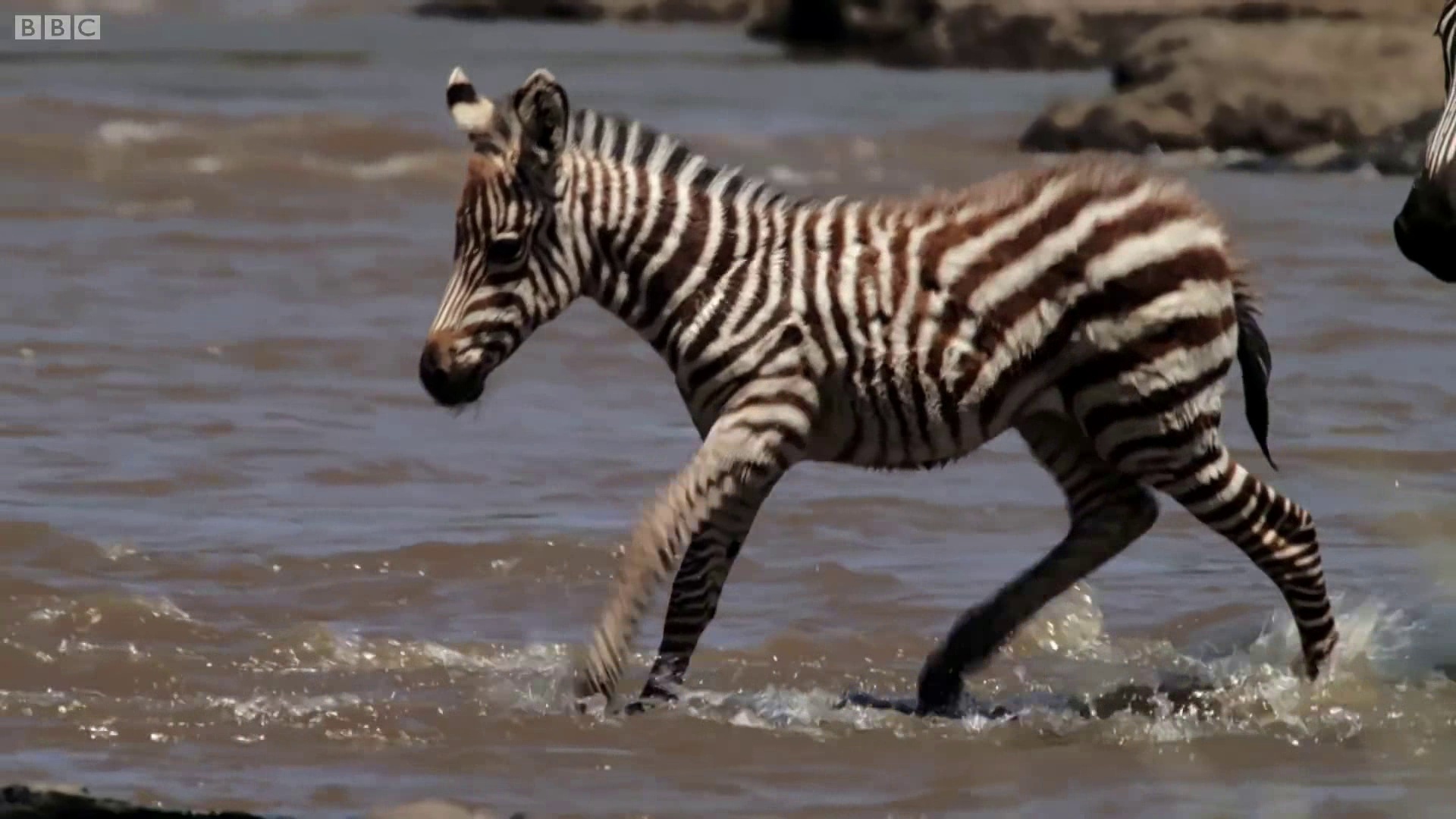}\hfill 

\vspace{3pt}

  \includegraphics[width=0.195\textwidth, height=2.8cm]{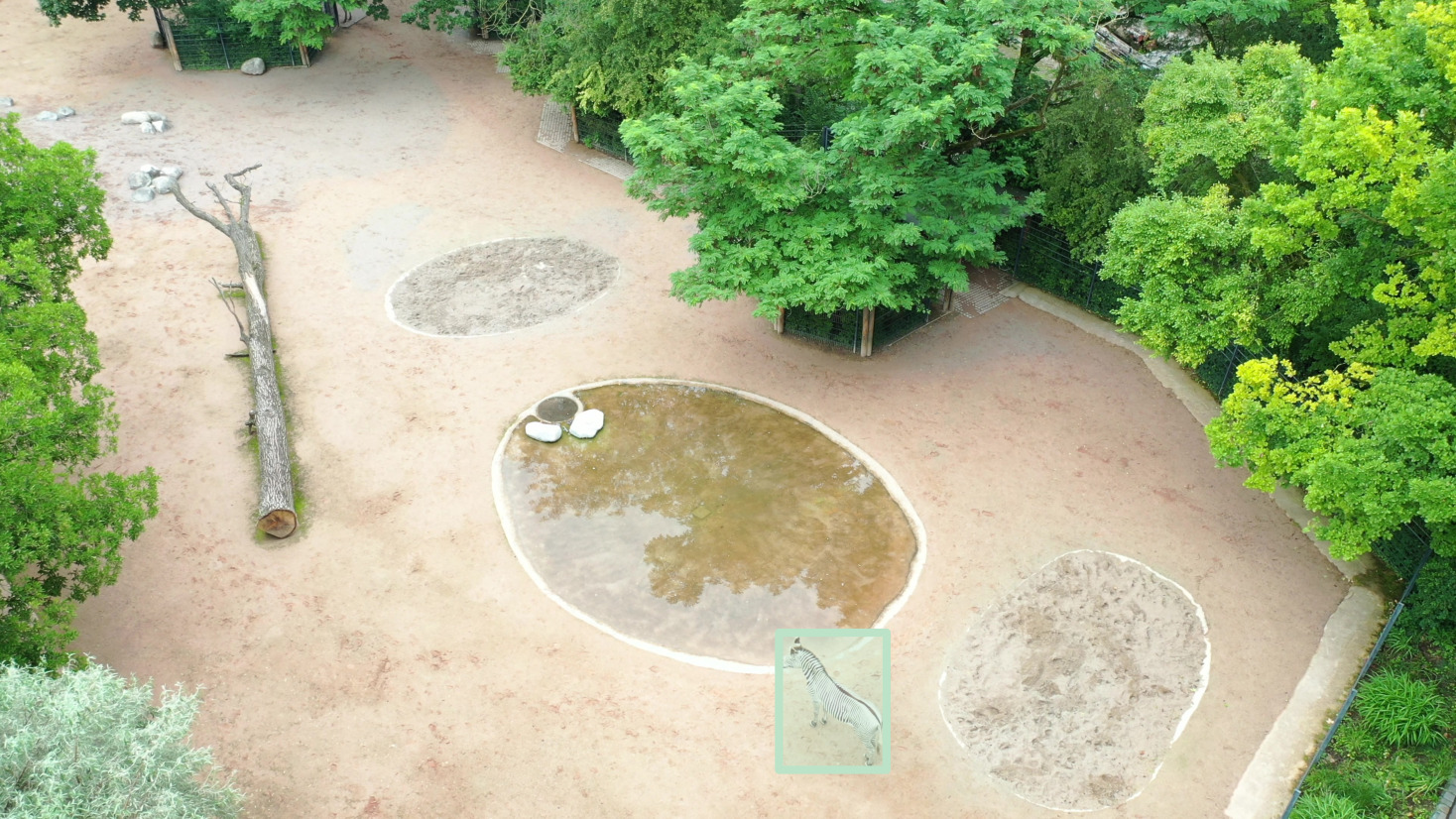}\hfill
  \includegraphics[width=0.195\textwidth, height=2.8cm]{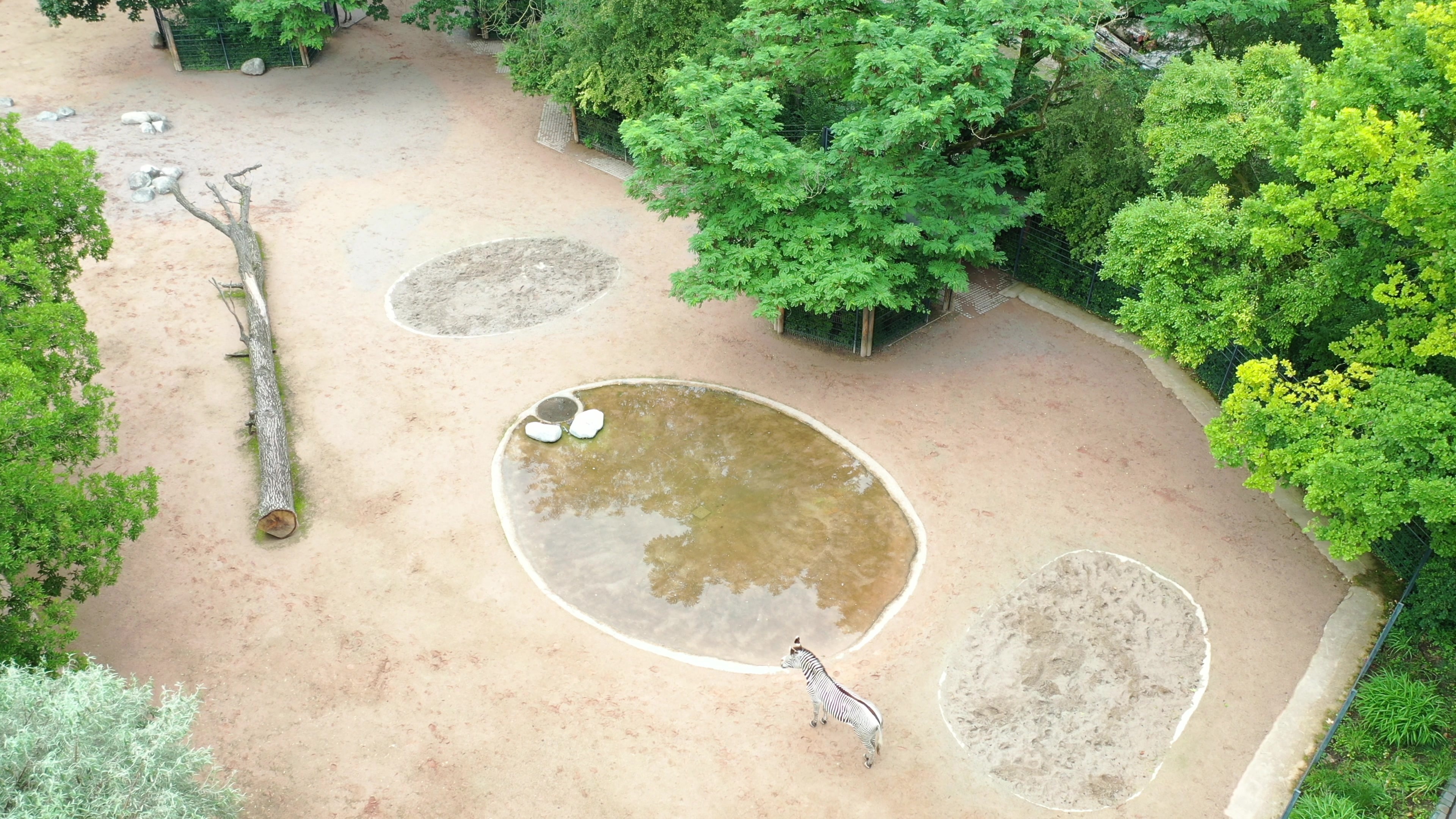}\hfill
  \includegraphics[width=0.195\textwidth, height=2.8cm]{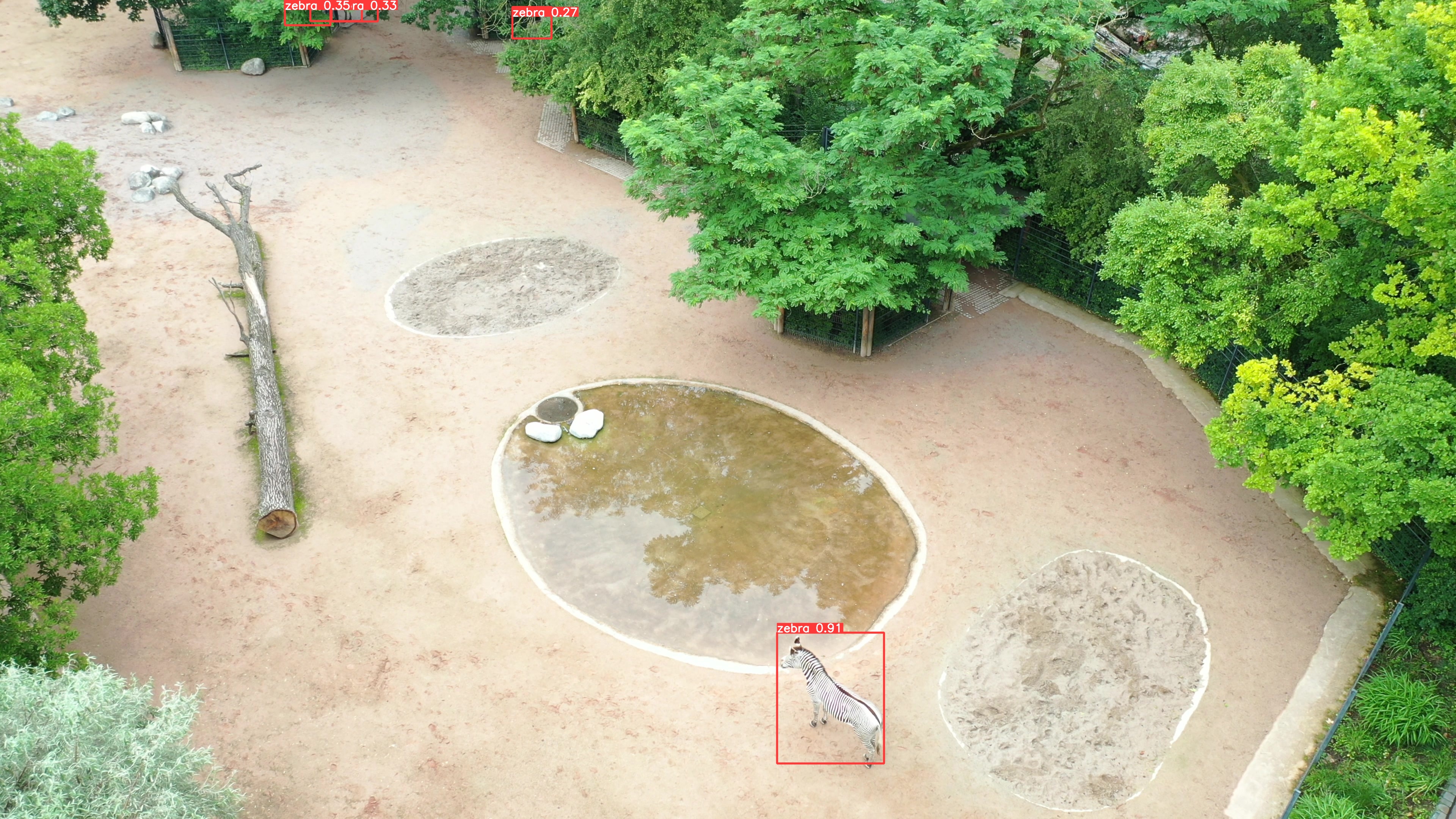}\hfill
  \includegraphics[width=0.195\textwidth, height=2.8cm]{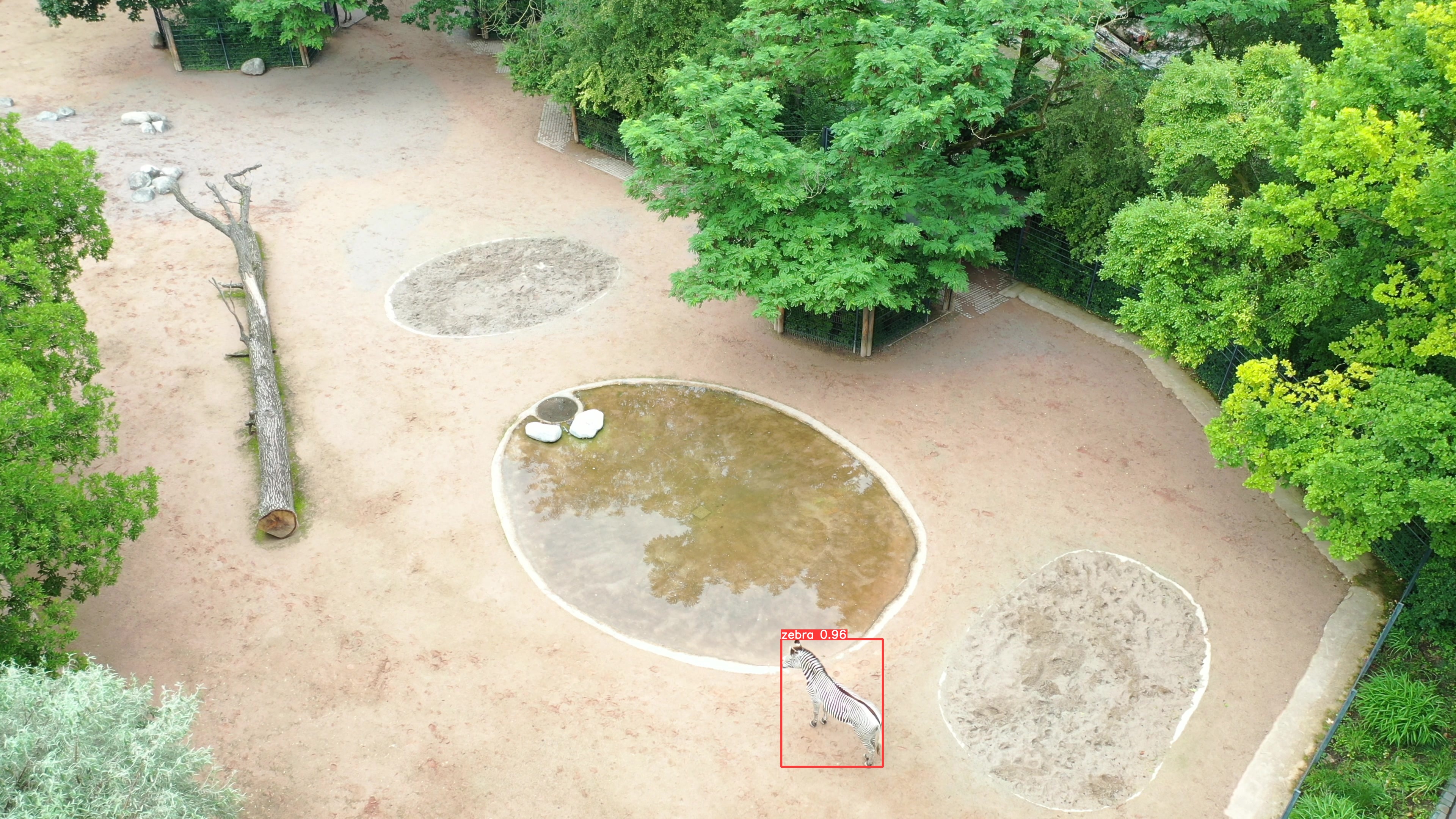}\hfill
  \includegraphics[width=0.195\textwidth, height=2.8cm]{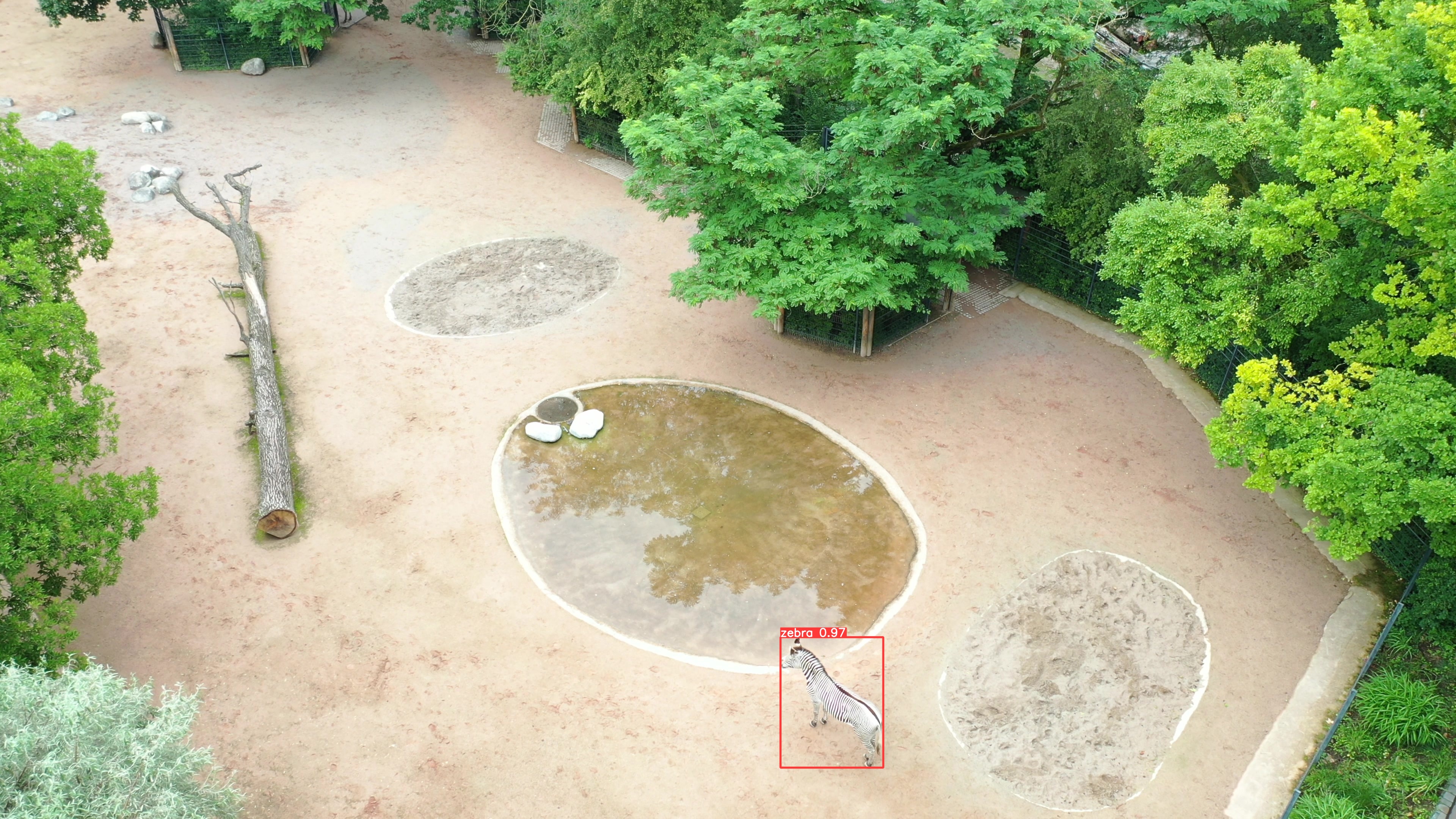}\hfill 

\vspace{3pt}

  \subcaptionbox{Ground-truth}{\includegraphics[width=0.195\textwidth, height=2.8cm]{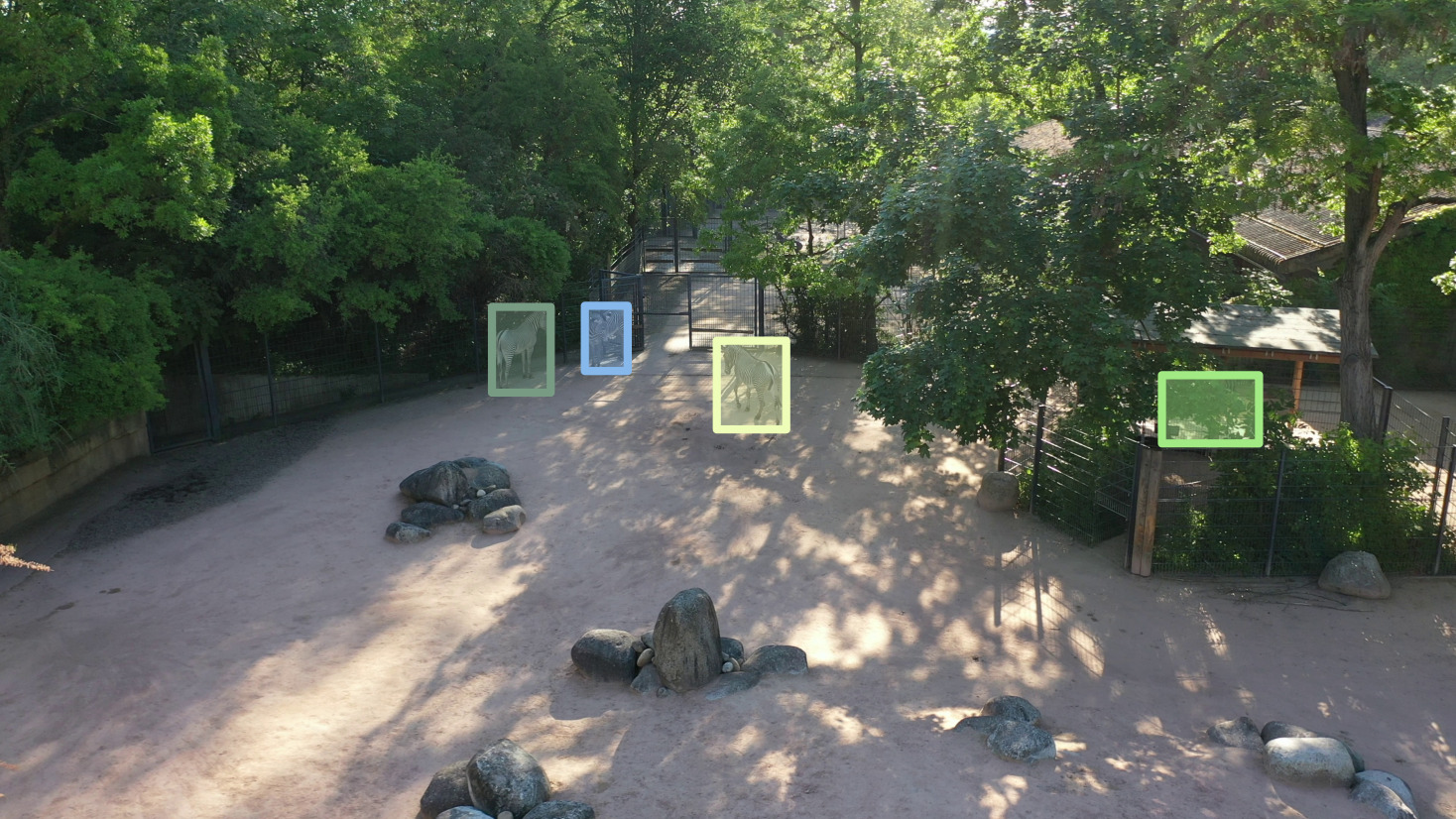}}\hfill
  \subcaptionbox{COCO}{\includegraphics[width=0.195\textwidth, height=2.8cm]{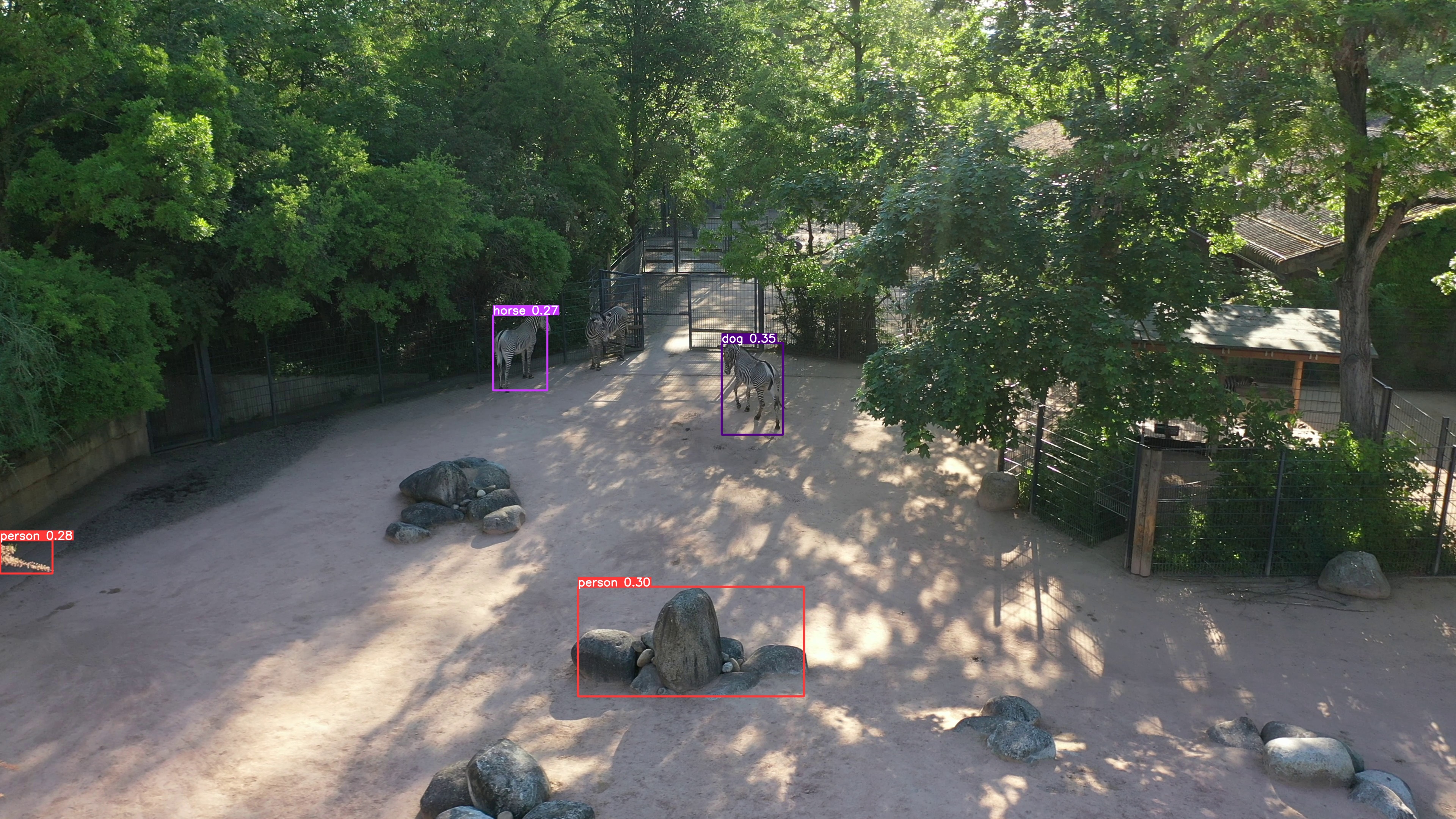}}\hfill
  \subcaptionbox{RP-1920}{\includegraphics[width=0.195\textwidth, height=2.8cm]{fig/RP/rp.jpg}}\hfill
  \subcaptionbox{SC-1920}{\includegraphics[width=0.195\textwidth, height=2.8cm]{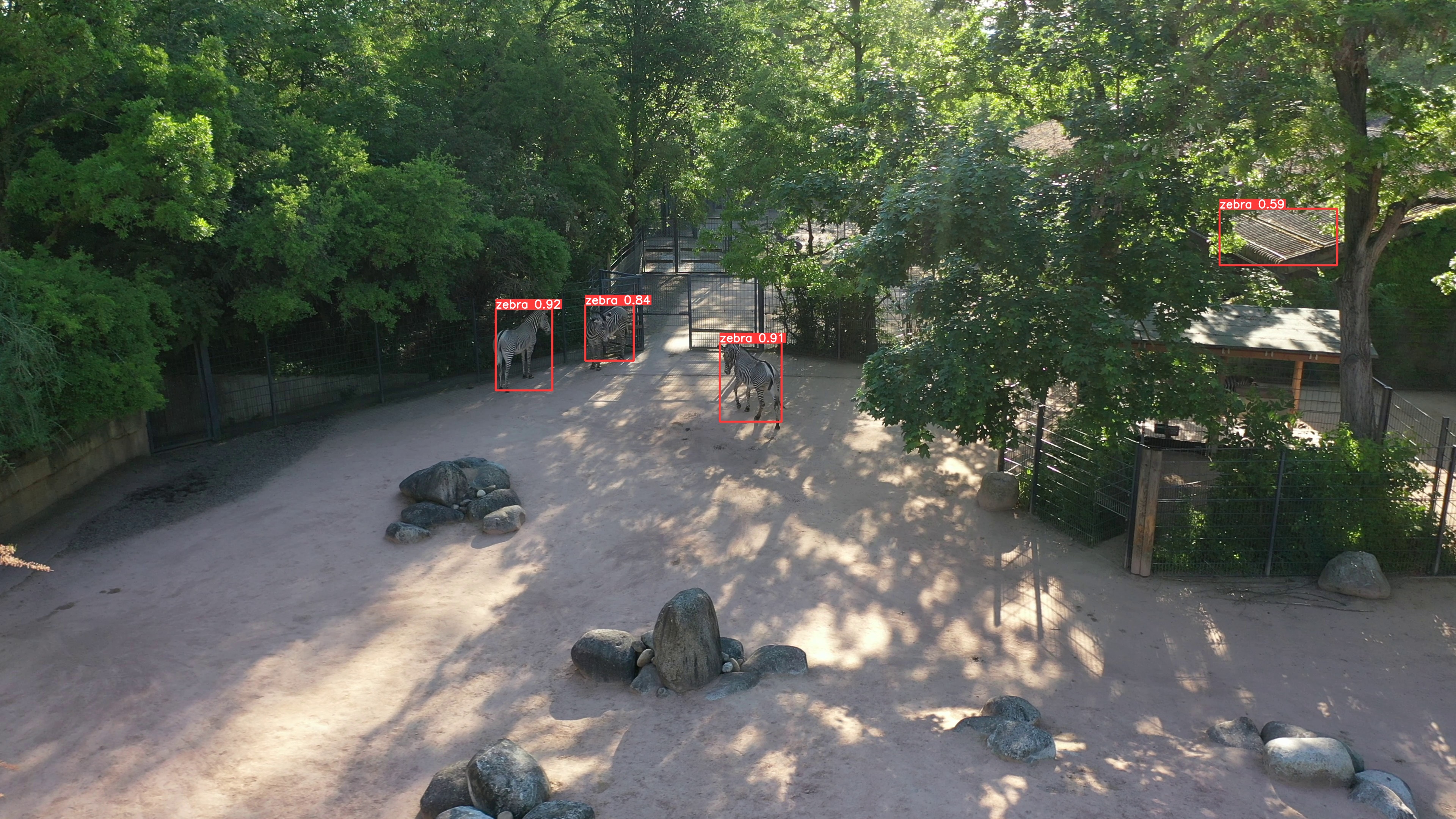}}\hfill
  \subcaptionbox{SC+COCO+$\text{R3}_{100}$-1920}{\includegraphics[width=0.195\textwidth, height=2.8cm]{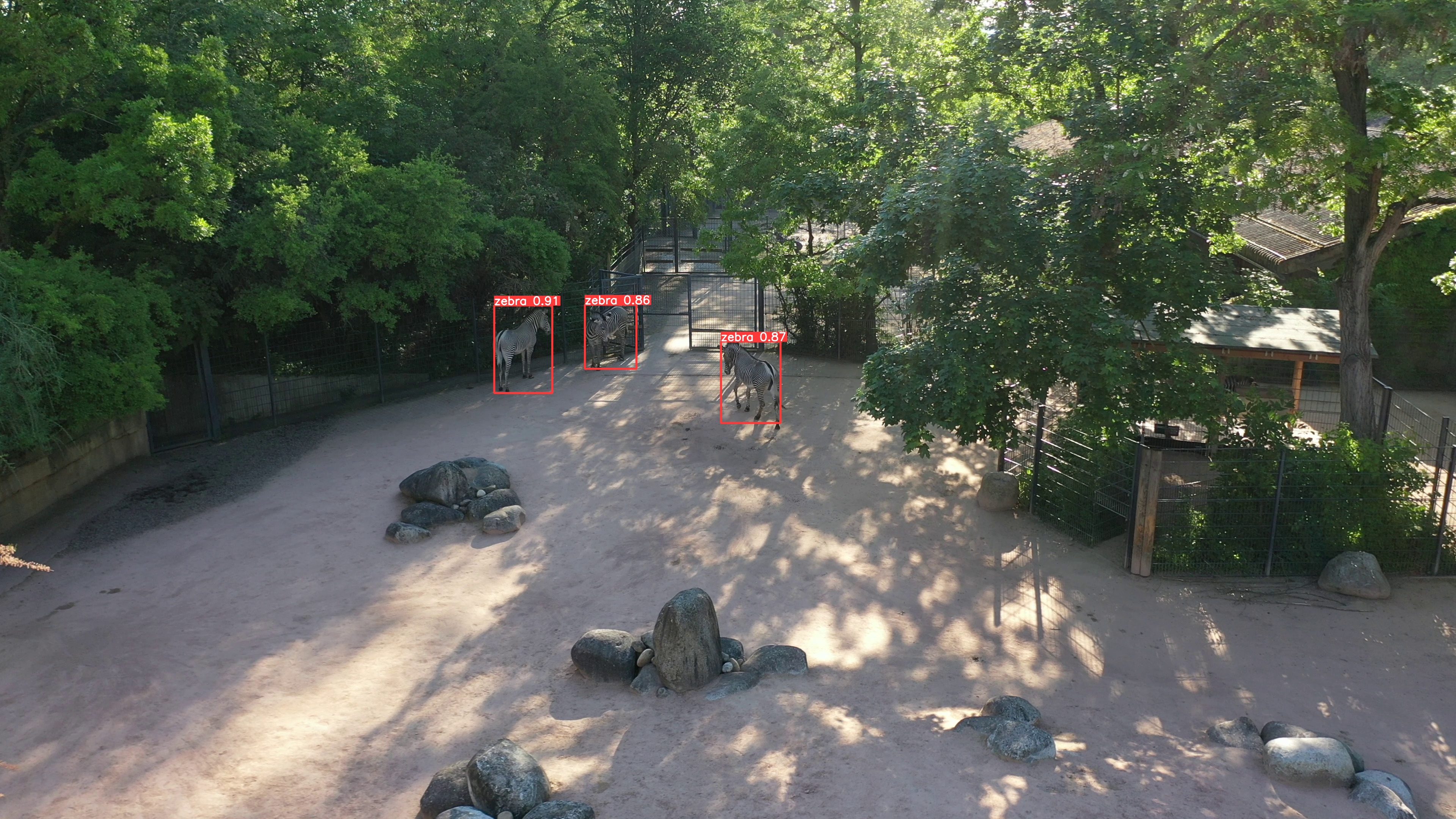}}\hfill 

\vspace{10pt}

  \caption{Sampled detections. We show in column (a) the ground truth and then the results obtained from (b) the default model (pre-trained on COCO), and the ones trained on (c) RP-1920, (d) SC-1920, and (e) SC+COCO+$\text{R3}_{100}$-1920. The images are randomly taken from the COCO (first row), APT-36K, R2, and RP (last row) datasets. Best viewed in color and zoomed-in.}
  \label{fig:labels-detections-results}
\end{figure*}

All our results are reported in Tab.~\ref{tab:res}. Additionally, we present randomly sampled images from the COCO, \mbox{APT-36K}, R2, and RP datasets for the main models in Fig.~\ref{fig:labels-detections-results}. Now, we proceed to analyse the results that we report in the table. First, we can notice that the models trained on the bigger image size show higher performances across all datasets and metrics. This is true both for synthetic, i.e. SF and SF-1920, SC and SC-1920, and the real data, i.e.  RP and RP-1920. The only exception is the model trained with RP which in the APT-36K performs $\sim10\%$ better than RP-1920.
Overall, the model pre-trained on the COCO dataset works well only on the APT-36K dataset with a mAP50 of $\sim88\%$, further showing the low variability of these datasets and the incapability to generalize to both different points of view or scenarios. Indeed, the YOLO model pre-trained on COCO achieves at most $\sim58\%$ accuracy on our data, with an overall weighted average of $\sim47\%$. The fact that the COCO data is representative of the APT-36K dataset can be evinced also by the performance obtained by the model trained with RP+COCO-1920 dataset.
Considering now the synthetic data, i.e. SF-1920 and SC-1920, we can see that the best model overall is SC-1920 which achieves $\sim5\%$ higher mAP and mAP50 across all tests, with a peak of $\sim15\%$ on the R3 dataset. This is probably related to the first of the two generation procedures, which resulted in long distances between the zebras and the cameras (see Sec.~\ref{sec:envdatagen}). Our real data instead comprises mostly zebras that are reasonably nearby the drone as seen from the pictures in Fig.~\ref{fig:labels-wil} and Fig.~\ref{fig:labels-detections-results} in the third and fourth rows. The synthetic models may perform poorly on RP validation set and APT-36K due to the generation process. These sets have diverse images, including zebras near the camera in a side view or hidden behind bushes and trees, e.g. second and fourth row in Fig.~\ref{fig:labels-detections-results}. Moreover, by comparing SC-1920 with the model pre-trained on COCO, we can see how, across all data excluding APT-36K, we obtain higher performances on both metrics of considerable amounts, ranging between $\sim20\%$ and $\sim45\%$.

We can now compare the differences between the models trained on synthetic data and real data. For this, we will focus on comparing SC-1920 and RP-1920. The weighted average gap is only $4.9\%$ in the mAP50 and $3.2\%$ on the mAP. The big difference in the simple average is mostly linked to the results obtained in the validation set of RP, which was to be expected. Indeed, we can notice how the model trained on synthetic data performs considerably worse in the RP dataset, with a $\sim58\%$ reduction in mAP50 and $\sim41\%$ on mAP. A similar result is depicted when we consider tests on the R3\_Dx data, with reductions of $\sim16\%$ and $\sim6-10\%$ for the two considered metrics. Nonetheless, with all other datasets, the model trained on synthetic data is comparable to the one trained on real-world captured images of just $1-5\%$. Recall that the RP model was trained on the RP dataset itself, composed of images from the Rx experiments and additional images from point-of-views not generated by our procedure. This clearly demonstrates that, on the considered datasets, the model trained solely with the synthetic data generated using the pipeline described above is perfectly capable of detecting zebras by achieving similar performance on all but two datasets when compared with RP, and significantly overcoming the model pre-trained on COCO in all but APT-36K dataset.

Finally, we consider the mixed models. Unsurprisingly, the one based only on real data, i.e. RP+COCO-1920, performs well on all datasets. The slight reduction in performance in the R2 and R3 datasets is well compensated by the generalization in the APT-36K. This is also the model with the highest average mAP and mAP50. We believe that this is mostly linked to how the dataset was built, with RP that contains data from all Rx experiments combined with the 1916 training images of COCO. Despite that, it is interesting to notice how the models trained with a mixture of synthetic and real data are capable of generalizing across all the datasets as well. Specifically, combining SC and COCO, i.e. SC+COCO-1920, resulted practically in a significant improvement of the performance solely in the APT-36K dataset. Minor improvements are noticeable in the other datasets as well, If to this we add 100 samples from R3, i.e. SC+COCO+$\text{R3}_{100}$-1920, we then achieve considerable improvements in the performance w.r.t. SC-1920 on all datasets. 

The most noticeable are the ones on APT-36K, of around 55\%, and on the RP dataset, of around 10\%. The improvement in the R3 is to be expected since we mixed 100 images from that set. Nonetheless, it is remarkable that just a small change in the data brought a $\sim34\%$ increase in the mAP for this validation test. The SC+COCO+$\text{R3}_{100}$-1920 is the model with the highest weighted average precisions and is the second best when considering the average mAP and mAP50. We believe that this model would be further improved by having more samples from the COCO dataset in the validation set or, overall, a better-balanced set of samples. Considering that SC is made of 18K images, and both COCO and $\text{R3}_{100}$ make up for 2K training images and only 300 validation ones, we can expect an `overfit' of the final selected model towards scenes which are strongly represented by the synthetic images. Also, in this case, the significant difference in the average mAP and mAP50 is mostly linked to the gap in the results in the RP validation set. For completeness, we also trained the SC+COCO+RP-1920 model, i.e. using the closeby synthetic data, the coco data, and the small set of real data which was precisely labelled. As expected, this is the model which performs best in the majority of the tests, excluding the APT-36K dataset, where the pretrained model performs best, and in R3\_D1. However, we must note that RP contains data from all R1, R2 and R3 datasets in both the training and validation sets. Thus, the results in these case are clearly driven by this information. What is interesting to notice is that all mixed models perform similarly in the APT-36K dataset, with $\sim 70\%$ of mAP50 and $\sim 38\%$ mAP, further indicating that a better balancing in the validation set might further boost the performance of these models. Alternatively, a more representative generation strategy could be employed, by including camera locations relative to the zebras more similar to the ones that we can find in the APT-36K or in the COCO dataset. The results suggest that such an approach would be effective as well, perhaps in conjunction with a minimal amount of annotated real data. Finally, considering that zebra stripes are notoriously specific to the individual, it is interesting to notice how, despite the fact we use the same texture for all our generated zebras, we are still able to generalize to different individuals well. This suggests that the network does not focus and learn specifically the pattern it is shown, but rather the general appearance of the animal itself.

\section{CONCLUSIONS}
\label{sec:conc}

In this work, we first demonstrated that the currently available datasets do not generalize well to the task of detecting zebras captured from an aerial point of view. To solve this, we generated a large-scale synthetic dataset of zebras by using GRADE, a state-of-the-art framework for synthetic data generation. The dataset, which is the first of its kind both in terms of size and visual realism, has been released for the benefit of the community. By using that, we performed extensive evaluations by training and testing YOLO with a wide range of combinations of real and synthetic data. This provides strong evidence that the visual realism of the data generated is very high, because our models showed performances which are as good as the one obtained by a detector trained on real-world labelled data alone. Using synthetic information we can surpass the process of collecting and labelling data in controlled scenarios, thus avoiding the probable introduction of errors. Further testing by using combined synthetic and a small amount of real data showed that we can successfully generalize to a wide variety of scenarios. A known limitation that we need to address is the realism of the adopted environments and more precise placement strategies, which we believe could solve both the generalization problem and the usage of high-resolution images by the network. Future works include the generation of videos instead of just static images, testing with different network architectures like SSD~\cite{ssd} or RCNN~\cite{maskrcnn}, and using the synthetic data for different tasks such as keypoints detection.
	\bibliographystyle{IEEEtran}

\end{document}